\pdfoutput=1  
\documentclass{article}

\PassOptionsToPackage{numbers,sort&compress}{natbib}
\usepackage[preprint]{neurips_2026}        

\usepackage[utf8]{inputenc}
\usepackage[T1]{fontenc}
\usepackage{hyperref}
\hypersetup{
  pdftitle={Learning Scene-Grounded Interaction Priors for Scene-Aware Human Motion Prediction},
  pdfauthor={Donghyeon Soon; Jiil Yoon; Hyeondong Kim; Daehee Park}
}
\usepackage{url}
\usepackage{graphicx}
\usepackage{amsmath}
\usepackage{amssymb}
\usepackage{amsfonts}
\usepackage{bm}
\usepackage[table]{xcolor}
\usepackage{booktabs}
\usepackage{multirow}
\usepackage{subcaption}
\usepackage{pifont}
\usepackage{nicefrac}
\usepackage{microtype}
\usepackage{placeins}   

\newcommand{\Tch}{\mathcal{T}}
\newcommand{\Stu}{\mathcal{S}}
\newcommand{\Lz}{\mathcal{L}_{z}^{\mathrm{KD}}}
\newcommand{\LY}{\mathcal{L}_{Y}^{\mathrm{KD}}}
\newcommand{\g}{\cellcolor{gray!15}}
\newcommand{\cmark}{\ding{51}}   
\newcommand{\xmark}{\ding{55}}

\newif\ifverify
\verifytrue      

\title{SGIP: Learning Scene-Grounded Interaction Priors for Scene-Aware Human Motion Prediction}

\author{%
  Donghyeon~Soon\textsuperscript{1}\thanks{Equal contribution.}\quad
  Jiil~Yoon\textsuperscript{1}\footnotemark[1]\quad
  Hyeondong~Kim\textsuperscript{1}\quad
  Daehee~Park\textsuperscript{1, 2}\thanks{Corresponding author.} \\
  \textsuperscript{1}DGIST \quad
  \textsuperscript{2}UNIVA \\
  \texttt{\{dhsoon,wldlf2819,hdkim,dhpark\}@dgist.ac.kr}
}

\begin{document}
\maketitle
\begin{figure}[h]
    \centering
    \includegraphics[width=\linewidth]{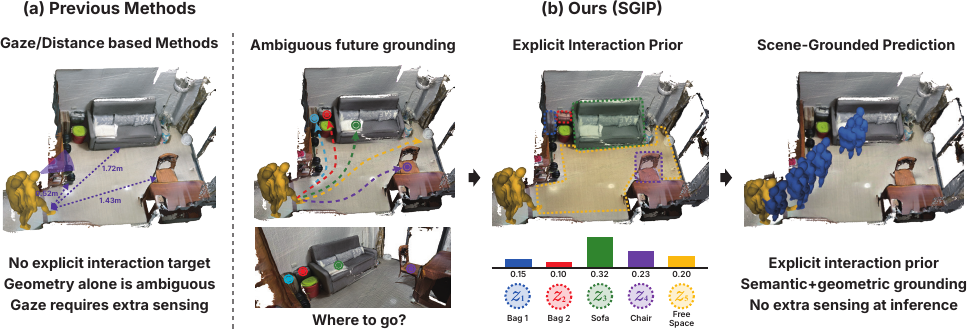}
    \vspace{-0.5em}
    \caption{
    \textbf{Overview of the proposed method.}
    Scene-aware human motion prediction (HMP) requires resolving \textit{future-motion ambiguity}.
    \textbf{(a)} Previous methods handle this ambiguity \textit{implicitly}, relying on motion--scene distance or gaze without an explicit interaction target; geometry alone is ambiguous when several objects lie at similar distances, and gaze requires extra sensing.
    \textbf{(b)} Given the same observed motion, a person may approach a bag, sit on the sofa or chair, or freely navigate the room.
    \textit{SGIP} resolves this by predicting an explicit \textit{scene-grounded interaction prior}, a distribution over candidate scene tokens covering both object instances and free space, from the observed motion and 3D scene alone.
    Distilled at training time from an intent-aware teacher that combines semantic and geometric scores, the prior places the highest probability on the sofa token, which conditions both trajectory and full-body pose prediction to correctly anticipate walking toward and sitting on the sofa, with no extra sensing at inference.
    }
    \label{fig:teaser}
    \vspace{-0.4em}
\end{figure}

\begin{abstract}
Human motion prediction (HMP) aims to forecast future movement from an observed motion sequence, and recent methods have incorporated 3D scene geometry and object semantics to improve long-term plausibility.
However, the destination of future motion remains ambiguous in cluttered scenes: the same observed motion can lead to different interaction targets or navigation modes.
Existing methods either handle this ambiguity implicitly or rely on auxiliary signals like gaze that are hard to obtain in practice.
To address this, we propose \textit{SGIP}, which explicitly predicts where in the scene future motion will be grounded across both objects and open navigation regions, and uses it to condition future trajectory and full-body pose prediction.
This scene-grounded interaction prior is learned via intent distillation: \textit{at training time only}, a teacher model scores candidate scene regions by combining an LLM prompted with a textual description of the person's intent with geometric motion-scene features.
A student model learns to reproduce these scores from only the observed motion and 3D scene, making \textit{intent and LLM unnecessary at test time}.
Across multiple benchmark datasets, SGIP outperforms prior scene-aware baselines on both trajectory and pose accuracy in a realistic deployment setting where neither gaze nor intent is available at inference, demonstrating the effectiveness of the proposed framework. The project page is available at \url{https://sgip2026.github.io}.
\end{abstract}
\section{Introduction}
\label{sec:introduction}
Human motion prediction (HMP) aims to forecast future human poses and trajectories from an observed motion sequence, and is essential for autonomous systems, embodied agents, and human--robot collaboration~\citep{gui2018teaching, mao2019ltd, lou2024sif3d}.
While early approaches largely modeled motion dynamics in isolation~\citep{martinez2017human, mao2019ltd}, recent HMP has shifted toward 3D scenes, where predictions must respect both the physical layout and the semantic affordances of the environment~\citep{hassan2021posa,wang2021longterm}.
These scene-aware methods improve long-term plausibility by incorporating human--object context, future contact, and gaze-based intention cues~\citep{corona2020context,mao2022contact,lou2024sif3d,yu2025vision}.

However, future motion in such scenes is inherently ambiguous: the same 
observed motion can correspond to multiple valid ways of interacting 
with or navigating through the scene.
For example, a person approaching a dining area may sit on one of several chairs or pass through the space toward another room; each possibility induces a different trajectory, body orientation, and contact pattern.
Prior work has attempted to resolve this ambiguity by relying on auxiliary cues such as gaze~\citep{lou2024sif3d, yu2025vision} or manually specified interaction targets~\citep{hassan2021samp, wang2022humanise}.
%
However, such cues require specialized sensors or per-frame supervision and are difficult to obtain in deployment. 
A practical HMP model should therefore make scene-aware predictions from the observed motion and 3D scene alone (Figure~\ref{fig:teaser}).

%

To address this, we propose SGIP, which makes scene-interaction reasoning an explicit, supervisable target by predicting a scene-grounded interaction prior, a distribution over candidate scene tokens covering both interactive objects and navigable regions.
SGIP predicts this prior from the observed motion and 3D scene alone, and uses it to condition future trajectory and full-body pose prediction.
However, learning the prior directly from observation is challenging since the observed motion alone often does not specify a clear interaction target.
We address this through \textit{privileged intent distillation}.
At training time, we leverage a textual description of the person's intent (e.g., ``take some rest''), which is practical to obtain through a vision-language model rather than requiring per-sample human annotation. 
A teacher prompts a large language model with this intent to score how well each candidate scene region aligns with the person's goal.
Yet textual intent alone is insufficient when a scene contains multiple instances of the same object class, where the LLM cannot distinguish which specific instance the person is heading toward.
We therefore combine the LLM's semantic score with a geometric compatibility score derived from motion--scene distance and orientation, which uses the observed motion to disambiguate between same-class instances.
Together, these form the teacher's intent-conditioned distribution over scene tokens.
At inference, where neither textual intent nor an LLM is available, a lightweight MLP-based student reproduces this distribution from motion and scene alone, inheriting the teacher's intent-aware scene-interaction reasoning through distillation.

Our contributions are summarized as follows:
(1) We introduce a \textit{scene-grounded interaction prior}, an explicit distribution over candidate scene tokens covering both object-centric interactions and free-space navigation, that turns scene-interaction reasoning into a supervisable target.
(2) We propose a \textit{privileged intent distillation} framework in which textual intent and an LLM supervise a teacher at training, while the student reproduces the resulting prior from motion and scene alone, requiring no language model or auxiliary cue at inference.
(3) On GIMO and PROX, the observation-only student consistently improves trajectory and pose accuracy over scene-aware baselines without auxiliary cues, with the privileged teacher additionally serving as a performance ceiling.
\section{Related Work}

\subsection{Human Motion Prediction}
Human motion prediction (HMP) forecasts future human poses or trajectories from observed motion 
and underpins applications such as autonomous driving, human--robot collaboration, 
and animation~\cite{fragkiadaki2015erd, martinez2017human, jain2016structural, mao2019ltd, alahi2016sociallstm, salzmann2020trajectronpp}.
Architectures span recurrent~\cite{fragkiadaki2015erd, martinez2017human}, 
graph-based~\cite{mao2019ltd, dang2021msrgcn, li2022spgsn, sofianos2021stsgcn}, 
and attention- or Transformer-based~\cite{aksan2021spatio, mao2020history, xu2023auxformer, xu2023eqmotion} 
formulations, with probabilistic variants (VAEs, flows, GANs, diffusion) 
modeling multi-future ambiguity~\cite{aliakbarian2020stochastic, yuan2020dlow, 
mao2021gsps, salzmann2022motron, barquero2023belfusion, chen2023humanmac}.
Yet motion-only predictors overlook the surrounding 3D environment and 
frequently violate scene constraints in deployment.
Scene-aware HMP addresses this by injecting scene context, evolving from 
2D image- or bounding-box-based formulations~\cite{cao2020long, corona2020context} 
to point-cloud fusion~\cite{zheng2022gimo, mao2022contact, huang2023diffusion}, typically built on PointNet-style backbones~\cite{qi2017pointnet, qi2017pointnetpp} with diverse scene-grounded predictors~\cite{xing2024mdp, lou2024dimop3d, gao2024mcld};
gaze further serves as a behavioral proxy for intent in cluttered 
scenes~\cite{zheng2022gimo, lou2024sif3d, yu2025vision}.
Recent methods derive intention features from gaze to drive separate
trajectory and pose branches~\cite{lin2026humo3d}, or model
human--human and human--scene interactions in dynamic
multi-person environments~\cite{sun2026humof} captured in datasets such as HIK~\cite{tanke2023hik} and HOI-M$^3$~\cite{zhang2024hoim3}.
A complementary line synthesizes scene-aware motion conditioned on
object affordances~\cite{hassan2021posa, wang2021longterm,
jiang2024trumans, kulkarni2024nifty, li2024chois}
or textual action descriptions and activity
labels~\cite{wang2022humanise, cen2024textscene, wang2025sims},
rather than forecasting motion from observed histories.


\subsection{Language Models for Scene and Motion Understanding}

Large language models (LLMs) have recently been used to 
inject commonsense knowledge into 3D scene and motion tasks.
Motion--language alignment models adapt LLMs to translate 
textual descriptions into human motion 
sequences~\cite{jiang2023motiongpt}, and scene-aware variants 
further ground the generated motion in 3D environments 
through textual scene or activity 
descriptions~\cite{yang2024llm, wang2022humanise, cen2024textscene, 
wang2025sims}.
Closely related, several methods route language conditions 
through explicit \emph{affordance} or \emph{intent} 
representations as intermediate structure between language 
and motion~\cite{wang2024affordmotion, ghosh2023imos, 
diomataris2024wandr}, reflecting a broader shift toward 
semantically structured intermediates rather than direct 
text-to-motion mapping.
Beyond generation, LLMs serve as reasoners for 3D scene and 
human--scene understanding by aligning object semantics with 
language~\cite{hong20233dllm, zhao2025hisgpt, huang2023embodied}, and as 
grounding modules that connect language goals to spatial 
affordances for embodied agents~\cite{ahn2022saycan}.
A complementary line of work treats LLMs as judges or 
rubric-based scorers~\cite{zheng2023judging}, where 
natural-language criteria elicit interpretable, score-level 
outputs over candidate options.
Despite this flexibility, deploying LLMs at inference incurs 
substantial computational and latency cost, which is poorly 
suited to real-time motion forecasting in deployment.

\subsection{Privileged Information and Knowledge Distillation}

A natural alternative is to use such expensive or 
hard-to-obtain supervision only at training time.
Learning with privileged information~\cite{vapnik2009new, 
lopez2015unifying} formalizes this setting by exploiting 
signals available during training but absent at inference, 
while knowledge 
distillation~\cite{hinton2015distilling, romero2015fitnets} 
transfers such information from a teacher to a more 
deployable student through output- or 
intermediate-feature-level imitation.
A concrete realization in vision is \emph{modality 
hallucination} or \emph{modality distillation}, where a 
privileged modality (e.g., depth or optical flow) supervises 
a student that operates from a single observable modality at 
inference~\cite{hoffman2016modality, garcia2018modality,
zhao2022pkd}; in human trajectory forecasting, multi-modal teachers have similarly been distilled into students that observe fewer modalities~\cite{jeong2025multimodal}.
In the context of human motion in 3D scenes, behavioral or 
semantic cues such as gaze~\cite{zheng2022gimo, lou2024sif3d, 
yu2025vision}, language 
goals~\cite{wang2022humanise, cen2024textscene}, or 
explicitly stated intent~\cite{ghosh2023imos} can 
disambiguate human--scene behavior but are rarely accessible 
in practical deployment, making them natural candidates for 
privileged supervision.
We adopt this paradigm with two privileged training-time 
signals---textual intent and an LLM-based rubric scorer---and 
distill the resulting \emph{scene-grounded interaction prior} 
into an observation-only student, rather than transferring 
gaze- or language-conditioned motion outputs as in prior 
distillation.

\begin{figure*}[t]
    \centering
    \includegraphics[width=\textwidth]{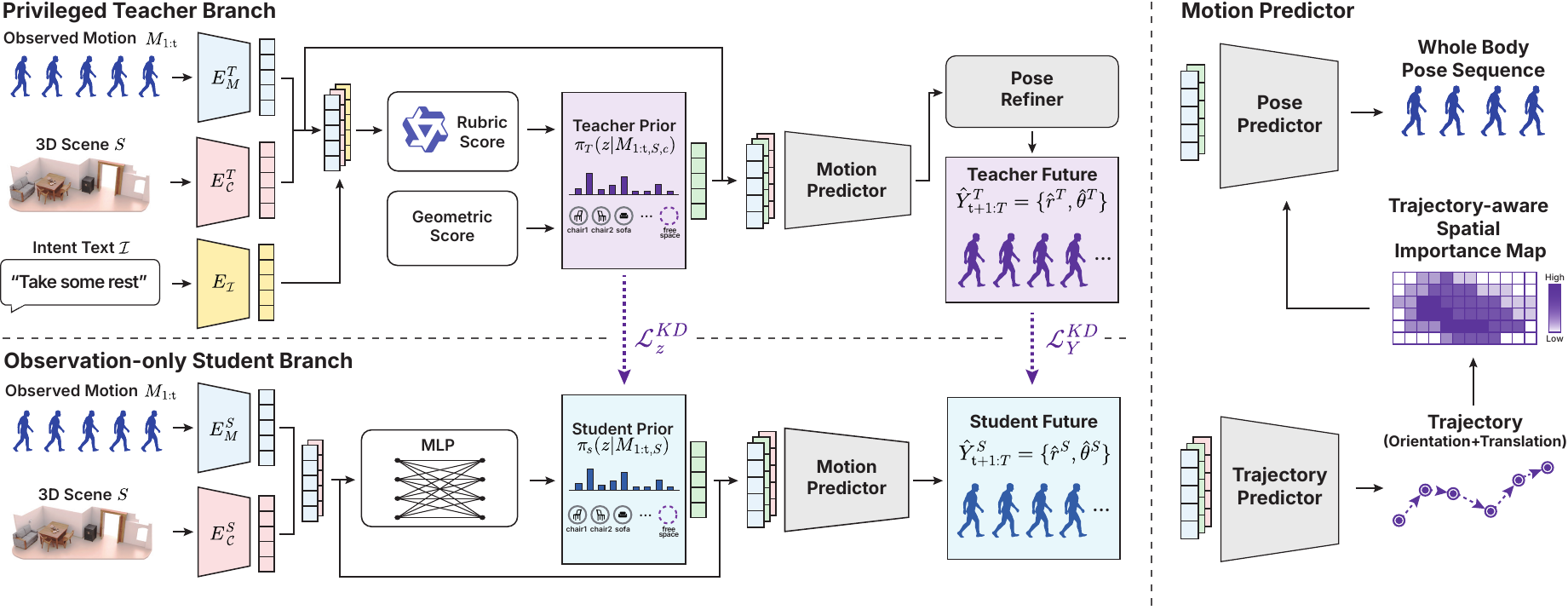}
    \caption{
        \textbf{Overall pipeline of \textit{SGIP}.} A privileged teacher branch 
        (top, training only) constructs an intent-conditioned prior $\pi^{\Tch}$ over 
        candidate grounding tokens by fusing an LLM-based rubric score 
        with a geometric compatibility score, with a Pose Refiner 
        further intent-aligning the teacher's pose. An observation-only 
        student branch (bottom) predicts $\pi^{\Stu}$ from motion $M_{1:t}$ 
        and scene $S$ alone via a scoring head $F_{\Stu}^{\text{grd}}$ 
        (a permutation-equivariant MLP), trained via prior distillation 
        $\Lz$ and motion-level distillation 
        $\LY$. The selected token then conditions the 
        motion predictor (right) that outputs future trajectory and 
        full-body pose.
    }
    \label{fig:pipeline}
\end{figure*}

\section{Proposed Method}
\label{sec:method}

SGIP is a two-branch teacher--student framework (Figure~\ref{fig:pipeline}).
The teacher fuses an LLM-based rubric score with a geometric compatibility score into an intent-conditioned interaction prior,
and a student approximates this prior from motion and scene alone, serving as the deployed model. Section~\ref{sec:problem} formalizes the problem;
Section~\ref{sec:prior} defines the scene-grounded interaction prior and its teacher and student forms;
Section~\ref{sec:decode} describes how the prior conditions trajectory and full-body pose prediction;
and Section~\ref{sec:distill} describes the distillation procedure.
Implementation details (VLM and LLM calls, hyperparameters, training schedule) are deferred to Appendices~\ref{app:supervision}--\ref{app:impl}.

\subsection{Problem Formulation}
\label{sec:problem}

We are given an observed motion sequence $M_{1:t} = \{x_1, \dots, x_t\}$,
where each frame $x_\tau = (\bm{r}_\tau, \bm{\theta}_\tau)$ consists of root motion $\bm{r}_\tau = (\bm{t}_\tau, \bm{R}_\tau)$,
with translation $\bm{t}_\tau \in \mathbb{R}^3$ and orientation $\bm{R}_\tau \in SO(3)$,
together with SMPL-X body pose $\bm{\theta}_\tau$~\cite{pavlakos2019smplx}.
The surrounding 3D environment is a point cloud $S \in \mathbb{R}^{N \times 3}$.
Our goal is to predict the future motion $\hat{Y}_{t+1:T} = \{\hat{x}_{t+1}, \dots, \hat{x}_T\}$ over a horizon $\Delta T = T - t$.
To make scene-interaction reasoning explicit, we extract a set of candidate \emph{grounding tokens} $\mathcal{C} = \mathcal{C}^{\text{obj}} \cup \mathcal{C}^{\text{nav}}$,
where $\mathcal{C}^{\text{obj}}$ contains object instances 
and $\mathcal{C}^{\text{nav}}$ is a single scene-global token representing object-agnostic navigable regions.
%
Since multiple tokens in $\mathcal{C}$ can afford plausible futures from the same observed motion,
we introduce a \emph{scene-grounded interaction prior} $\pi$ over $\mathcal{C}$:
\begin{equation}
\pi_i = p(z = i \mid M_{1:t}, S), \quad i \in \mathcal{C},
\label{eq:prior_def}
\end{equation}
where the latent $z$ indexes the token on which future motion is grounded.

At training time, we obtain a textual intent $\mathcal{I}$
(e.g., \textit{``take some rest''}) by applying a frozen
vision-language model (VLM) to training videos and corresponding
3D motion visualizations
(Appendix~\ref{sec:appendix-intent-extraction}).
The intent and an LLM-based rubric scorer construct a
privileged teacher prior $\pi^{\Tch}$, which is
distilled into the observation-only student prior
$\pi^{\Stu}$.
%
The most probable token under $\pi^{\Stu}$ conditions the motion predictor,
augmented at training with prior- and motion-level distillation losses against the teacher.
At inference, $\mathcal{I}$, the VLM, and the LLM are discarded.

\subsection{Scene-Grounded Interaction Prior}
\label{sec:prior}

We model future-motion grounding as a categorical distribution $\pi$ over the candidate set $\mathcal{C}$,
and construct two views of this distribution:
a privileged \emph{teacher} prior $\pi^{\Tch}$ that exploits intent and an LLM at training time,
and an observation-only \emph{student} prior $\pi^{\Stu}$ that approximates $\pi^{\Tch}$ from motion and scene alone at inference.

\paragraph{Token representations.}
Object instances $\mathcal{C}^{\text{obj}}$ are extracted via zero-shot 3D instance segmentation~\cite{huang2024openins3d} on $S$.
%
%
%
We retain object candidates that intersect a $120^\circ$ forward-facing cone whose axis bisects the observed head- and torso-facing directions, aggregated over the observation window.
This pruning uses only the observed motion (no gaze) and is shared by the teacher, the student, and all SGIP ablations (Appendix~\ref{app:cone}).

Each token $c_i = (\bm{\phi}_i, \bm{e}_i, \bm{m}_i)$ consists of three representations that capture complementary aspects of the token:
\begin{align}
\bm{\phi}_i &= [\bm{\phi}_{\text{geo}}(c_i),\, \bm{\phi}_{\text{rel}}(c_i, M_{1:t})]
    && \text{(geometric descriptor)}, \\
\bm{e}_i &= E_{\mathcal{C}}(c_i)
    && \text{(token embedding)}, \\
\bm{m}_i &\subseteq S
    && \text{(point-cloud mask).}
\end{align}
$\bm{\phi}_i$ captures where the token is and how the observed motion approaches it, which identifies the specific instance among same-class candidates being headed toward; $\bm{e}_i$ is a learned semantic representation of the interaction the token affords;
and $\bm{m}_i$ provides its 3D shape,
grounding physically plausible body configurations at the target.
The observed motion sequence is encoded into a single embedding 
$\bm{h}_M^b = E_M^b(M_{1:t}) \in \mathbb{R}^{d_M}$ for branch $b\in\{\Tch,\Stu\}$, where $E_M^b$ denotes the branch's motion encoder.

\paragraph{Teacher prior: combining semantic and geometric evidence.}
At training, the teacher constructs $\pi^{\Tch}$ by combining two complementary scores.
A \emph{semantic relevance score} obtained from an LLM rates each candidate against the textual intent $\mathcal{I}$:
\begin{equation}
s_i^{\text{rubric}} = \mathrm{LLM}(\mathcal{I}, c_i), \quad i \in \mathcal{C},
\label{eq:rubric}
\end{equation}
where the LLM is queried with a fixed prompt scoring two criteria---intent affordance (how well a candidate semantically supports the intent) and intent usability (how easily the intent can be realized using that candidate, given the scene). The rubric, LLM call, and output validation are given in Appendix~\ref{app:llm}, and the intent taxonomies in Appendix~\ref{sec:intent-taxonomy}.

A \emph{geometric compatibility score} aggregates fixed directional alignments between the observed motion and each candidate (details in Appendix~\ref{app:geo_score}):
\begin{equation}
s_i^{\text{geo}} = w_{\text{face}}\, \mathrm{align}_{\text{face}}(\bm{\phi}_i, M_{1:t}) + w_{\text{body}}\, \mathrm{align}_{\text{body}}(\bm{\phi}_i, M_{1:t}) + w_{\text{line}}\, \mathrm{align}_{\text{line}}(\bm{\phi}_i, M_{1:t}).
\label{eq:geo}
\end{equation}

The two scores are complementary.
The LLM captures intent--class affordance (\textit{a chair affords sitting}) but cannot disambiguate among multiple instances of the same class within a scene;
the geometric score grounds the semantic prior in the observed motion, identifying \emph{which} specific instance is being approached.
We fuse them via temperature softmax with fusion weight $\alpha$ and temperature $\tau$:
\begin{equation}
\pi_i^{\Tch} = \frac{\exp\bigl((s_i^{\text{rubric}} + \alpha\, s_i^{\text{geo}}) / \tau\bigr)}{\sum_{j \in \mathcal{C}} \exp\bigl((s_j^{\text{rubric}} + \alpha\, s_j^{\text{geo}}) / \tau\bigr)}.
\label{eq:teacher_prior}
\end{equation}
This yields an intent-conditioned prior that the student is trained to match.

\paragraph{Student prior: observation-only approximation.}
At inference, neither $\mathcal{I}$ nor the LLM is available, so the student 
reproduces $\pi$ from motion and the geometric token descriptors alone through 
a learnable scoring head $F_{\Stu}^{\text{grd}}$:
\begin{equation}
\pi_i^{\Stu}
  = \frac{\exp F_{\Stu}^{\text{grd}}(\bm{h}_M^{\Stu}, \bm{\phi}_i)}
         {\sum_{j\in\mathcal{C}} \exp F_{\Stu}^{\text{grd}}(\bm{h}_M^{\Stu}, \bm{\phi}_j)}.
\label{eq:student_softmax}
\end{equation}
Because $\mathcal{C}$ varies in size across scenes and tokens have no canonical order,
$F_{\Stu}^{\text{grd}}$ is implemented as a permutation-equivariant MLP over candidates~\cite{zaheer2017deepsets}.
The student is trained to match $\pi^{\Tch}$ via knowledge distillation (Section~\ref{sec:distill}),
inheriting the teacher's intent-aware grounding without any auxiliary signal at inference.

\subsection{Prior-Conditioned Motion Prediction}
\label{sec:decode}
Given the prior $\pi^b$ from branch $b \in \{\Tch, \Stu\}$,
the motion predictor $D^b$ produces future motion
$\hat{Y}_{t+1:T}^b = \{\hat{\bm{r}}_{t+1:T}^b, \hat{\bm{\theta}}_{t+1:T}^b\}$
conditioned on the most probable grounding token,
\begin{equation}
i^*_b = \arg\max_{i \in \mathcal{C}} \pi_i^b,
\label{eq:select}
\end{equation}
whose embedding $\bm{e}_{i^*_b}$ and point-cloud mask $\bm{m}_{i^*_b}$ are read off from the selected token $c_{i^*_b}$.
For multi-future prediction, we separately decode the
top-$K$ tokens ranked by $\pi^{\mathcal{S}}$ while keeping
the observed motion, scene, and model parameters fixed.
Only the selected token's embedding and mask change,
yielding $K$ token-conditioned future motions without
stochastic sampling or privileged inputs.
As shown on the right side of Figure~\ref{fig:pipeline}, $D^b$ follows a two-stage trajectory-then-pose paradigm~\cite{lou2024sif3d}.

\paragraph{Trajectory branch.}
The trajectory branch first encodes the 3D scene $S$ into per-point
features via a scene encoder $E_S$, then cross-attends the motion
embedding $\bm{h}_M^b$ to these features while conditioning on the selected token's embedding
$\bm{e}_{i^*_b}$, which steers the predicted path toward the chosen interaction target
(Appendix~\ref{app:arch:decoder}):
\begin{equation}
\hat{\bm{r}}_{t+1:T}^b = D_{\text{traj}}^b(\bm{h}_M^b, S, \bm{e}_{i^*_b}).
\label{eq:traj}
\end{equation}
The predicted trajectory then induces a spatial importance map $P^b$
aggregating path proximity, forward-facing alignment, and proximity to the selected token's region:
\begin{equation}
P^b = f_{\text{spatial}}(\hat{\bm{r}}_{t+1:T}^b, S, \bm{m}_{i^*_b}).
\label{eq:spatial}
\end{equation}
\paragraph{Pose branch.}
$P^b$ is added as a bias to the pose branch's scene-attention logits,
keeping body pose consistent with the planned path and interaction target:
\begin{equation}
\tilde{\bm{\theta}}_{t+1:T}^b = D_{\text{pose}}^b(\bm{h}_M^b, \hat{\bm{r}}_{t+1:T}^b, S, P^b).
\label{eq:pose_branch}
\end{equation}
Architectural details ($f_{\text{spatial}}$, encoders, attention blocks) are in Appendix~\ref{app:arch}.
For the teacher only, a pose refiner $\mathrm{Refine}$ that leverages $\mathcal{I}$ is applied
to recover intent-conditioned pose semantics beyond motion reconstruction
(e.g., distinguishing sitting from standing in front of the same chair);
details are in Appendix~\ref{app:refiner}.

\subsection{Privileged Intent Distillation}
\label{sec:distill}

To deploy the teacher's intent-aware reasoning at inference, we distill it into the student via two complementary signals:
\emph{prior distillation} aligns \emph{where} future motion is grounded ($\pi^{\Tch} \to \pi^{\Stu}$),
and \emph{motion-level distillation} aligns \emph{how} the model moves toward that target ($\hat{Y}^{\Tch} \to \hat{Y}^{\Stu}$).

\paragraph{Prior distillation.}
We align $\pi^{\Stu}$ with $\pi^{\Tch}$ via forward KL divergence:
\begin{equation}
\Lz = D_{\mathrm{KL}}\bigl(\mathrm{sg}[\pi^{\Tch}] \,\|\, \pi^{\Stu}\bigr)
= \sum_{i \in \mathcal{C}} \pi_i^{\Tch} \log \frac{\pi_i^{\Tch}}{\pi_i^{\Stu}},
\label{eq:lz_kd}
\end{equation}
where $\mathrm{sg}[\cdot]$ stops gradients to $\pi^{\Tch}$.
The mass-covering forward direction forces the student to cover every token the teacher considers,
preserving the full intent-conditioned grounding structure rather than only its top-1 and preventing mode collapse.

\paragraph{Motion-level distillation.}
We distill the teacher's motion via $L_1$ loss on trajectory and pose:
\begin{equation}
\LY
  = \lambda_{\text{traj}}
    \bigl\|\hat{\bm{r}}_{t+1:T}^{\Stu}
    - \mathrm{sg}[\hat{\bm{r}}_{t+1:T}^{\Tch}]\bigr\|_1
  + \lambda_{\text{pose}}
    \bigl\|\tilde{\bm{\theta}}_{t+1:T}^{\Stu}
    - \mathrm{sg}[\hat{\bm{\theta}}_{t+1:T}^{\Tch}]\bigr\|_1,
\label{eq:ly_kd}
\end{equation}
where $\hat{\bm{\theta}}_{t+1:T}^{\Tch}$ is the refined teacher pose,
and $\lambda_{\text{traj}}, \lambda_{\text{pose}}$ balance the two terms.
Crucially, since the teacher pose is refined with intent, the intent-conditioned pose semantics (e.g., sitting versus standing in front of the same chair) can be distilled into the student.

\paragraph{Training schedule.}
The teacher is trained first with its motion loss $\mathcal{L}_{\text{mot}}^{\Tch}$ and additional Refine-specific terms (Appendix~\ref{app:impl}).
The student then freezes the entire teacher and optimizes
\begin{equation}
\mathcal{L}^{\Stu} = \mathcal{L}_{\text{mot}}^{\Stu} + \lambda_z\, \Lz + \lambda_Y\, \LY,
\label{eq:student_obj}
\end{equation}
where $\mathcal{L}_{\text{mot}}^{\Stu}$ shares the form of $\mathcal{L}_{\text{mot}}^{\Tch}$.
\section{Experiments}
\label{sec:experiments}

\begin{table*}[t]
\centering
\caption{
Main comparison on GIMO and PROX. The ``Auxiliary Cue'' column denotes
auxiliary signals beyond $M+S$: $g$=gaze, $\mathcal{I}$=textual intent.
$\dagger$: diagnostic privileged branch on GIMO (not deployed);
$*$: our gaze-free re-evaluation of gaze-using baselines;
$\ddagger$: gaze-free variant reported in the original paper.
SGIP results on GIMO are mean$\pm$std over 5 seeds.
\textbf{Bold}/\underline{underline}: best/second-best within block.
}
\label{tab:main}
\small
\setlength{\tabcolsep}{3pt}
\renewcommand{\arraystretch}{0.92}
\begin{tabular}{m{1.2em}|lccccc}
\toprule
& \textbf{Method}
& \textbf{Auxiliary Cue}
& \textbf{Traj-path} $\downarrow$
& \textbf{Traj-dest} $\downarrow$
& \textbf{MPJPE-path} $\downarrow$
& \textbf{MPJPE-dest} $\downarrow$ \\
\midrule
\multirow[c]{18}{*}{\rotatebox[origin=c]{90}{\textbf{GIMO}}}
& \multicolumn{6}{c}{\textit{Models with additional or privileged cues}} \\
\cmidrule{2-7}
& LTD~\cite{mao2019ltd} w/ gaze             & $g$ & 655 & 801 & 155.6 & 202.0 \\
& SPGSN~\cite{li2022spgsn} w/ gaze          & $g$ & 702 & 982 & 156.0 & 202.0 \\
& AuxFormer~\cite{xu2023auxformer} w/ gaze  & $g$ & 622 & 746 & 161.5 & 201.1 \\
& BiFu~\cite{zheng2022gimo}                 & $g$ & 661 & 727 & 167.8 & 205.0 \\
& SIF3D~\cite{lou2024sif3d}                 & $g$ & 590 & 666 & 156.6 & 195.7 \\
& GAP3DS~\cite{yu2025vision}                & $g$ & 575 & \underline{623} & \textbf{141.2} & \textbf{171.4} \\
& HuMo3D~\cite{lin2026humo3d}               & $g$ & 543 & 651 & \underline{146.1} & \underline{185.5} \\
& \g$\textit{SGIP}_{\mathrm{priv}}^\dagger$ & \g$g$ & \g\underline{528} & \g635 & \g149.9 & \g190.6 \\
& \g$\textit{SGIP}_{\mathrm{priv}}^\dagger$ & \g$\mathcal{I}$ & \g\textbf{485} & \g\textbf{556} & \g151.2 & \g187.8 \\
\cmidrule{2-7}
& \multicolumn{6}{c}{\textit{Observation-only models}} \\
\cmidrule{2-7}
& LTD~\cite{mao2019ltd}                     & -- & 673 & 890 & \underline{156.4} & 203.0 \\
& SPGSN~\cite{li2022spgsn}                  & -- & 700 & 893 & 157.0 & 202.2 \\
& AuxFormer~\cite{xu2023auxformer}          & -- & 732 & 902 & 165.1 & 202.4 \\
& SIF3D$^{*}$~\cite{lou2024sif3d}           & -- & 700 & 823 & 164.2 & \underline{199.6} \\
& GAP3DS$^{*}$~\cite{yu2025vision}          & -- & \underline{642} & \underline{701} & 170.1 & 206.4 \\
& HuMo3D$^{\ddagger}$~\cite{lin2026humo3d}  & -- & \underline{642} & 731 & 178.3 & 210.1 \\
& \g\textit{SGIP} & \g--
& \g\textbf{549}\,{\scriptsize($\pm$5.0)}
& \g\textbf{681}\,{\scriptsize($\pm$7.6)}
& \g\textbf{153.1}\,{\scriptsize($\pm$1.8)}
& \g\textbf{195.1}\,{\scriptsize($\pm$2.6)} \\
\midrule
\multirow[c]{5}{*}{\rotatebox[origin=c]{90}{\textbf{PROX}}}
& LTD~\cite{mao2019ltd}               & -- & 677 & 753 & 223.4 & 226.6 \\
& SPGSN~\cite{li2022spgsn}            & -- & 595 & 663 & 286.3 & 271.3 \\
& AuxFormer~\cite{xu2023auxformer}    & -- & 553 & 727 & \underline{199.2} & 240.8 \\
& SIF3D$^{*}$~\cite{lou2024sif3d}     & -- & \underline{488} & \underline{659} & 228.5 & \underline{217.4} \\
& \g\textit{SGIP} & \g-- & \g\textbf{470} & \g\textbf{576} & \g\textbf{194.6} & \g\textbf{197.7} \\
\bottomrule
\end{tabular}
\end{table*}

\subsection{Experimental Setup}

\textbf{Datasets.}
We evaluate \textit{SGIP} on GIMO~\cite{zheng2022gimo}, a real-world indoor dataset with full-body SMPL-X motion and reconstructed 3D scenes, following the standard scene-aware HMP protocol~\cite{zheng2022gimo,lou2024sif3d,yu2025vision}: the model observes a motion prefix in a 3D scene and predicts the future global trajectory and full-body pose.
We additionally evaluate on PROX~\cite{hassan2019prox}, which has no gaze annotations, so gaze-using baselines are evaluated without gaze.
For both datasets, training-time pseudo-intents are extracted offline by a frozen VLM with dataset-specific taxonomies (Appendix~\ref{sec:appendix-intent-extraction}), and SGIP is trained and evaluated separately on each.
We further compare SGIP in multi-human, dynamic scenes on HIK~\cite{tanke2023hik} and HOI-M$^3$~\cite{zhang2024hoim3} following the HUMOF protocol~\cite{sun2026humof}; these results are reported in Appendix~\ref{app:interactive}.

\textbf{Baselines.} \emph{Pose-only} baselines (LTD~\cite{mao2019ltd}, SPGSN~\cite{li2022spgsn}, AuxFormer~\cite{xu2023auxformer}) forecast from body dynamics alone. \emph{Scene- and gaze-aware} baselines (BiFu~\cite{zheng2022gimo}, SIF3D~\cite{lou2024sif3d}, GAP3DS~\cite{yu2025vision}, HuMo3D~\cite{lin2026humo3d}) condition on the scene with gaze; to match our deployment assumption, we also report their gaze-free variants.
We additionally compare SGIP with a conditional diffusion-based motion generation model adapted to HMP in Appendix~\ref{app:diffusion}.
We report the deployed student as \textit{SGIP} and the privileged teacher as $\textit{SGIP}^{\dagger}_{\mathrm{priv}}$ (diagnostic only); in the privileged variants, intent ($s^{\text{rubric}}_i+\alpha s^{\text{geo}}_i$) or gaze (proximity and alignment of each candidate to the final-frame 3D gaze point) enters only through the candidate scores, so all variants share the same predictor (Appendix~\ref{app:cues}).

\textbf{Metrics.} \emph{Traj-path}/\emph{Traj-dest} measure global trajectory error averaged over the horizon and at the final frame, and \emph{MPJPE-path}/\emph{MPJPE-dest} the mean per-joint position error~\cite{ionescu2013human36m, mao2019ltd} on the same horizons~\cite{zheng2022gimo,lou2024sif3d,yu2025vision}.
For multi-future evaluation, we report minADE@$K$/minFDE@$K$ over $K$ predictions, FDE@minADE (the final error of the prediction with minimum ADE), and the average pairwise distance (APD) for diversity.

\textbf{Implementation.} Object tokens are obtained via OpenIns3D~\cite{huang2024openins3d}; intents are extracted with Qwen3-VL-8B~\cite{bai2025qwen3vl} and rubric scores with Qwen3-14B~\cite{yang2025qwen3}, both run once offline and cached.
After a one-time preprocessing of a static scene (61.7\,s), the student runs at 132.8\,ms per observation (7.5\,FPS) on one RTX~3090, below the 500\,ms observation interval of GIMO (Appendix~\ref{app:runtime}).

\subsection{Main Results}
\label{sec:main-results}

Among observation-only models, \textit{SGIP} achieves the best result on all four metrics on both GIMO and PROX (Table~\ref{tab:main}), indicating that the gains do not stem from GIMO-specific tuning.
Against the strongest observation-only baseline in each metric, SGIP reduces Traj-path by 93\,mm on GIMO (549 vs.\ 642\,mm) and Traj-dest by 83\,mm on PROX (576 vs.\ 659\,mm).
Pose-only methods exhibit large trajectory errors (673--732\,mm Traj-path on GIMO), and scene-aware baselines still trail \textit{SGIP}: scene-level conditioning alone does not resolve \emph{which} object the future motion engages, and gaze encodes \emph{where the subject looks} rather than \emph{what the subject intends to do}.
We attribute the gains to (1) the \textbf{explicit grounding prior}, whose selected token provides a spatial mask and semantic embedding that anchor the predictor to a specific target, yielding the largest gains on trajectory metrics where grounding ambiguity dominates; and (2) \textbf{privileged intent distillation}, which transfers intent-aware grounding and motion structure to the student, with $\textit{SGIP}^{\dagger}_{\mathrm{priv}}$ serving as a supervision upper bound rather than a deployable model.

\textbf{Privileged cues.} With the same predictor, intent yields lower trajectory error than gaze (485/556 vs.\ 528/635\,mm), indicating that knowing \emph{what} the person intends is more informative for grounding than knowing where they look.
GAP3DS remains stronger in privileged pose error (141.2/171.4 vs.\ 149.9/190.6\,mm with gaze) because it feeds an explicitly generated interaction pose to its decoder, whereas SGIP's cue only determines \emph{where} the interaction occurs (Appendix~\ref{app:cues}).

\begin{table}[t]
\centering
\caption{Multi-future evaluation on GIMO. SGIP separately decodes its top-$K$ grounding tokens; root-trajectory errors and APD are in meters (Appendix~\ref{app:diffusion}).}
\label{tab:multifuture}
\small
\setlength{\tabcolsep}{8pt}
\begin{tabular}{ccccc}
\toprule
$K$ & APD $\uparrow$ & minADE@$K$ $\downarrow$ & minFDE@$K$ $\downarrow$ & FDE@minADE $\downarrow$ \\
\midrule
1 & --    & 0.549 & 0.681 & 0.681 \\
2 & 0.884 & 0.517 & 0.601 & 0.605 \\
\rowcolor{gray!15}
3 & \textbf{0.927} & \textbf{0.467} & \textbf{0.523} & \textbf{0.571} \\
\bottomrule
\end{tabular}
\end{table}

\begin{figure*}[t]
\centering
\includegraphics[width=0.74\linewidth]{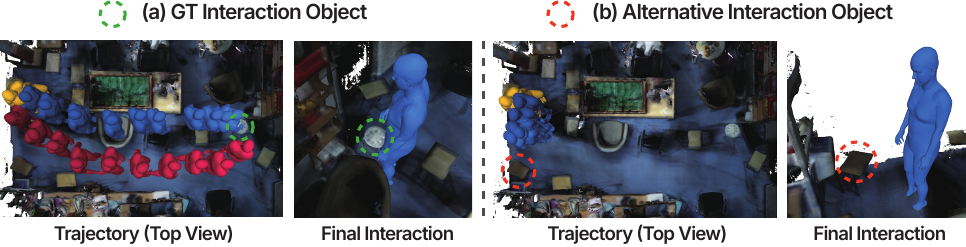}
\caption{
\textbf{Multimodal futures from $\pi^{\Stu}$.}
We condition the same fixed student predictor $D^{\Stu}$ on two
different grounding tokens drawn from $\pi^{\Stu}$:
\textbf{(a)} the ground-truth interaction object
(\textcolor{green!60!black}{green} dashed circle), and
\textbf{(b)} an alternative high-probability token of $\pi^{\Stu}$
(\textcolor{red}{red} dashed circle).
Each panel visualizes the resulting predicted trajectory
(top-down view) and the final interaction pose; only the
selected token $(\bm{m}_i, \bm{e}_i)$ fed into $D^{\Stu}$ varies.
}
\label{fig:multimodal}
\end{figure*}

\subsection{Multi-Future Evaluation}
\label{sec:multifuture}

We decode the top-$K$ tokens of $\pi^{\Stu}$ separately, $K\in\{1,2,3\}$, with the observed motion, scene, and model parameters fixed.
Increasing $K$ from 1 to 3 reduces minADE/minFDE from 0.549/0.681 to 0.467/0.523\,m and FDE@minADE from 0.681 to 0.571\,m, while APD reaches 0.927\,m (Table~\ref{tab:multifuture}): lower-ranked tokens provide distinct alternatives, some closer to the ground truth than the top-1, and each alternative is associated with a ranked, interpretable scene token rather than obtained by stochastic sampling.
Because FDE@minADE also decreases, the alternative that is closest on average also ends closer to the ground-truth destination, so the gain does not merely come from spreading predictions over more of the scene.
As Figure~\ref{fig:multimodal} illustrates, conditioning $D^{\Stu}$ on another high-probability token yields a distinct, scene-consistent trajectory and arrival pose, i.e., $D^{\Stu}$ has learned a token-conditional motion distribution.
A comparison with a conditional diffusion model under the same budget of three predictions is reported in Appendix~\ref{app:diffusion}.

\subsection{Ablation Studies}
\label{sec:ablation}

\begin{table}[t]
\centering
\caption{
Ablations on GIMO (5 seeds). \textbf{(a)} Distillation components $\Lz$ (prior) and $\LY$ (motion). \textbf{(b)} Hard vs.\ soft top-$k$ token aggregation at inference with the same trained student. \textbf{(c)} Top-1 accuracy (\%) on the GIMO-annotated target for teacher-side priors ($R$: rubric, $G$: geometry) and for the same student head distilled from different targets; bold marks the best within each group.
}
\label{tab:ablation_combined}
\begin{minipage}[t]{0.48\textwidth}
\centering
\subcaption{Distillation components.}
\label{tab:ablation_student}
{\small
\setlength{\tabcolsep}{4.5pt}%
\begin{tabular}{cc cc cc}
\toprule
\multicolumn{2}{c}{KD}
  & \multicolumn{2}{c}{Traj $\downarrow$}
  & \multicolumn{2}{c}{MPJPE $\downarrow$} \\
\cmidrule(lr){1-2} \cmidrule(lr){3-4} \cmidrule(lr){5-6}
$z$ & $Y$ & path & dest & path & dest \\
\midrule
              &              & 740 & 838 & 169.1 & 204.8 \\
              & $\checkmark$ & 856 & 896 & 164.1 & 198.9 \\
$\checkmark$  &              & \underline{574} & \underline{688} & \textbf{151.8} & \underline{198.6} \\
\rowcolor{gray!15}
$\checkmark$  & $\checkmark$ & \textbf{549} & \textbf{681} & \underline{153.1} & \textbf{195.1} \\
\bottomrule
\end{tabular}%
}
\end{minipage}\hfill
\begin{minipage}[t]{0.48\textwidth}
\centering
\subcaption{Hard vs.\ soft top-$k$.}
\label{tab:hard_vs_soft}
{\small
\setlength{\tabcolsep}{4.5pt}%
\begin{tabular}{l cc cc}
\toprule
\textbf{Aggregation}
  & \multicolumn{2}{c}{Traj $\downarrow$}
  & \multicolumn{2}{c}{MPJPE $\downarrow$} \\
\cmidrule(lr){2-3} \cmidrule(lr){4-5}
& path & dest & path & dest \\
\midrule
\rowcolor{gray!15}
Hard (top-1) & \textbf{549} & \textbf{681} & \textbf{153.1} & \textbf{195.1} \\
Soft (top-2) & 584          & 736          & 158.3          & 210.3 \\
Soft (top-3) & 715          & 877          & 176.5          & 210.2 \\
\bottomrule
\end{tabular}%
}
\end{minipage}

\vspace{0.6em}
\begin{minipage}{\textwidth}
\centering
\subcaption{Grounding accuracy (Acc@1, \%).}
\label{tab:grounding_acc}
\small
\setlength{\tabcolsep}{5pt}
\begin{tabular}{l ccc c cccc}
\toprule
& \multicolumn{3}{c}{Teacher-side prior} & & \multicolumn{4}{c}{Student (identical head), distillation target} \\
\cmidrule(lr){2-4} \cmidrule(lr){6-9}
& $R$ & $G$ & $G{+}R$ ($\pi^{\Tch}$) & & None & $\mathrm{KL}(R)$ & $\mathrm{KL}(G)$ & $\mathrm{KL}(G{+}R)$ \\
\midrule
Acc@1 & 54 & 51 & \textbf{86} & & 41 & 48 & 51 & \textbf{76} \\
\bottomrule
\end{tabular}
\end{minipage}
\end{table}

\paragraph{Distillation components.}
Without distillation, the student performs poorly on trajectory (Table~\ref{tab:ablation_student}), confirming that scene conditioning alone does not resolve grounding ambiguity; prior distillation $\Lz$ yields the largest single-component gain, reducing Traj-path/dest from 740/838 to 574/688\,mm.
Applying $\LY$ alone \emph{degrades} trajectory accuracy below the no-distillation baseline, which we attribute to \emph{grounding misalignment}: the teacher's output is conditioned on its token $i^{*}_{\Tch}$, while a student without prior supervision selects an inconsistent $i^{*}_{\Stu}$, so imitating the teacher's motion pulls predictions toward incompatible targets.
Consistently, the two branches agree on only 43\% of Top-1 tokens under $\LY$ alone versus 70\% once $\Lz$ is added (Appendix~\ref{app:agreement}). Prior distillation is thus a prerequisite for output-level distillation, which then further reduces trajectory error.

\paragraph{Hard vs.\ soft top-$k$ aggregation.}
SGIP conditions the predictor on the single token selected by the argmax of $\pi^{\Stu}$ (Eq.~\ref{eq:select}).
A natural alternative is to aggregate the top-$k$ tokens softly: their masks and embeddings are averaged with the renormalized probabilities $\tilde{\pi}^{\Stu}_i$, i.e., $\bar{\bm{m}}=\sum_{i}\tilde{\pi}^{\Stu}_i\bm{m}_i$ and $\bar{\bm{e}}=\sum_{i}\tilde{\pi}^{\Stu}_i\bm{e}_i$ over the top-$k$ set, and the same trained predictor consumes $(\bar{\bm{m}},\bar{\bm{e}})$ in place of $(\bm{m}_{i^*_{\Stu}},\bm{e}_{i^*_{\Stu}})$ (Appendix~\ref{app:soft-selector}).
As shown in Table~\ref{tab:hard_vs_soft}, soft aggregation degrades all four metrics, and the degradation grows with $k$: Traj-path/dest increase from 549/681 to 584/736\,mm with $k=2$ and to 715/877\,mm with $k=3$, and MPJPE-dest rises from 195.1 to about 210\,mm.
When the top-ranked tokens correspond to spatially separated targets (e.g., two chairs on opposite sides of a table), their weighted mixture describes no physically existing target: the averaged embedding steers the trajectory branch toward a location between the candidates, and the mixed mask spreads the spatial importance map $P^{b}$ (Eq.~\ref{eq:spatial}) over several disjoint regions, so the pose branch is no longer anchored to a single contact surface.
Moreover, since the predictor is trained with hard tokens, the mixed conditioning lies outside the distribution it has seen during training.
This mode-averaging behavior is precisely what explicit, discrete grounding is designed to avoid; the ambiguity among the top-ranked tokens is better exploited by decoding them separately as alternative futures (Section~\ref{sec:multifuture}) than by blending them into a single prediction.

\paragraph{Attribution of grounding accuracy.}
On the GIMO-annotated target (used only for evaluation), fusing rubric and geometry (86\%) clearly exceeds either alone (54\%/51\%; Table~\ref{tab:grounding_acc}).
To separate intent-conditioned supervision from the learnable head and candidate preprocessing, we train the same student head with identical inputs, candidates, and architecture, varying only the distillation target: it reaches 41\% without prior distillation and 51\% when distilling the geometry-only prior, but 76\% when distilling the fused, intent-conditioned prior.
This 25-point gap under an identical architecture isolates the contribution of intent-conditioned supervision (Appendix~\ref{app:grounding-acc}).

\begin{figure*}[t]
    \centering
    \includegraphics[width=\textwidth]{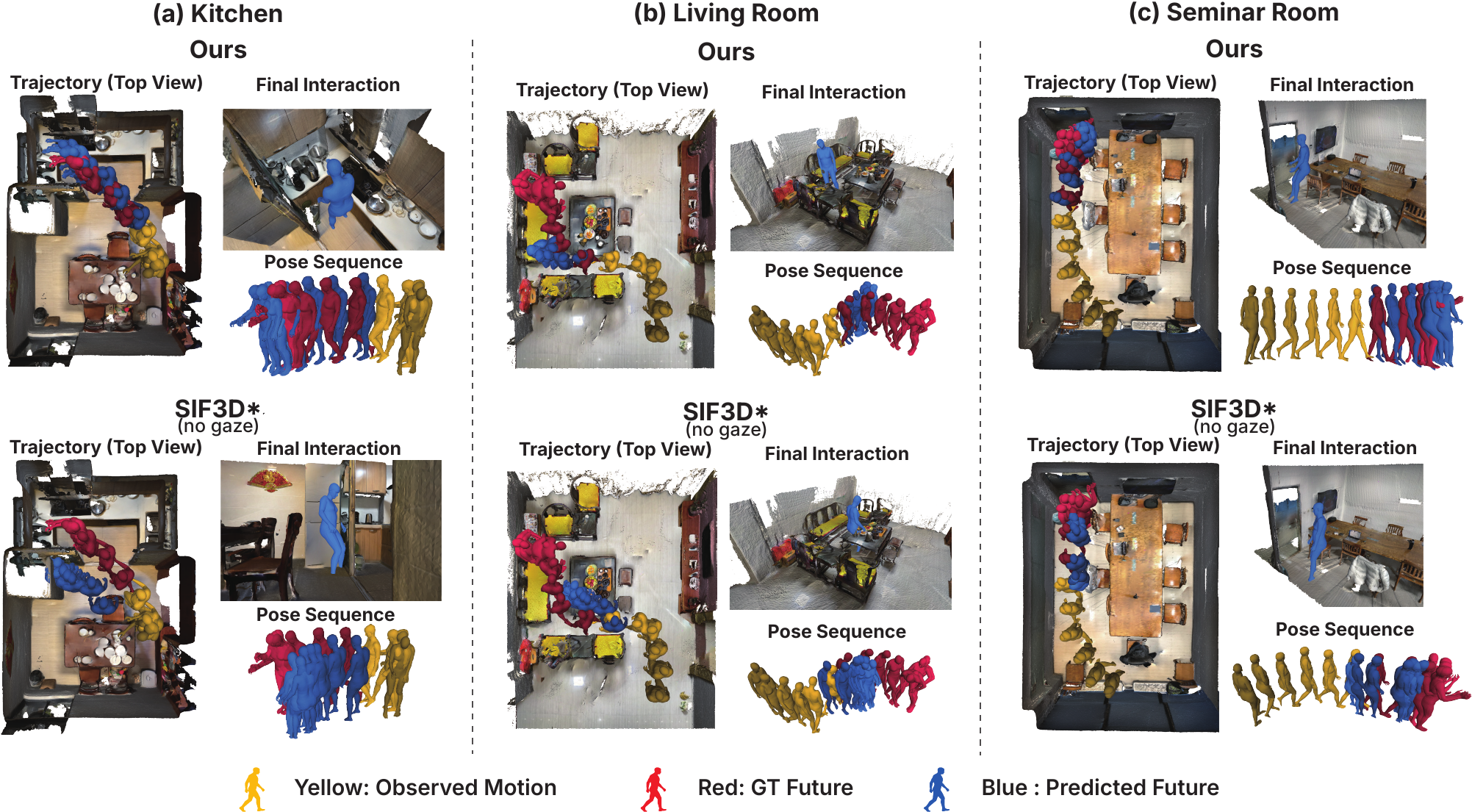}
    \caption{
    Qualitative comparison on three GIMO scenes
    (Kitchen, Living Room, Seminar Room). For each scene, the
    top row shows \textit{SGIP} (ours) and the bottom row shows
    the gaze-free baseline SIF3D$^{*}$. Each row visualizes the
    predicted trajectory (top-down view), the final interaction
    frame, and the full-body pose sequence.
    Yellow: observed motion; red: GT future; blue: predicted
    future.
    }
    \label{fig:qualitative}
\end{figure*}

\subsection{Qualitative Results}
Figure~\ref{fig:qualitative} compares \textit{SGIP} with the gaze-free SIF3D$^{*}$ on three cluttered GIMO scenes.
In the kitchen, SGIP follows the ground-truth path toward the counter, whereas SIF3D$^{*}$ turns toward the refrigerator side and ends at a different location.
In the living room, SIF3D$^{*}$ stalls near its starting position between the table and chairs, while SGIP advances along the ground-truth direction toward the sofa.
In the seminar room, both methods move along the wall, but SIF3D$^{*}$ falls short of the destination, whereas SGIP reaches the region where the ground-truth motion ends.
The gap is largest when several regions could plausibly attract the future motion: by committing to an explicit grounding token, SGIP produces a trajectory and arrival pose consistent with a single target instead of drifting between candidates, mirroring the trajectory gains in Table~\ref{tab:main}.
Moreover, even when its prediction departs from the recorded motion, SGIP commits to a different but reachable target and completes a coherent interaction with it, yielding an intent-consistent alternative rather than an averaged or stalled motion (Appendix~\ref{app:alternative}).
Appendix~\ref{app:additional-qualitative} also compares the SGIP teacher and student with SIF3D (with and without gaze); videos are in the supplementary material.

\section{Conclusion and Limitations}
\label{sec:conclusion}

We presented \textit{SGIP}, which makes human motion prediction explicitly grounded: it predicts where in the scene the future motion will arrive---at an object instance or in an open navigation region---and what it will interact with, and conditions trajectory and full-body pose prediction on this grounding.
The grounding is distilled from an LLM-guided, intent-aware teacher only at training time, so the deployed model requires neither intent annotations nor an LLM at inference.
In this sense, SGIP connects language-scale semantics to physical behavior: commonsense knowledge about what a space affords is transferred into a lightweight predictor that operates on the observed motion and 3D scene alone.
On GIMO and PROX, SGIP outperforms scene-aware HMP baselines on both trajectory and pose accuracy without gaze or intent at inference, and provides accurate, interpretable alternative futures.
We believe this matters beyond forecasting: robots that move in human living spaces must understand what can be done where, and must retain this understanding in deployment, where no language annotation is available; scene-grounded predictions of this kind could also turn recorded human motion into a useful teacher for learning humanoid full-body locomotion.

\textbf{Limitations.}
SGIP depends on the quality of candidate scene tokens (Appendix~\ref{app:noise}), requires a one-time instance segmentation for each new scene (Appendix~\ref{app:runtime}), and represents neighboring humans as static tokens, which limits accuracy in multi-human scenes (Appendix~\ref{app:interactive}).
Modeling time-varying human and object states and sampling futures directly from the full predicted prior are natural next steps.

\begin{ack}
This work was supported by the National Research Foundation of Korea (NRF) grants funded by the Korean government (MSIT) (No. RS-2026-25494733 and No. RS-2026-25561904, Vision-Language-Model-Based Adaptive Long-Horizon Task Planning for Autonomous Multi-Humanoid Collaboration, Contribution Rate: 20\%), the Institute of Information \& Communications Technology Planning \& Evaluation (IITP) grant funded by the Korean government (MSIT) (No. RS-2025-25442149, LG AI STAR Talent Development Program for Leading Large-Scale Generative AI Models in the Physical AI Domain), the Basic Science Research Program through the National Research Foundation of Korea (NRF) funded by the Ministry of Education (No. RS-2025-25420118), and the Culture, Sports and Tourism R\&D Program through the Korea Creative Content Agency (KOCCA) funded by the Ministry of Culture, Sports and Tourism in 2026 (No. RS-2026-25519568, Development of High-Precision Hand Technology and Service Technology Based on Physical AI, Contribution Rate: 10\%).

\par\medskip
\noindent\textbf{Competing Interests.}
The authors declare that they have no competing interests to disclose.
\end{ack}


\bibliographystyle{plainnat}
\bibliography{references}
\newpage
\appendix

\section{Overview and Information Availability}
\label{app:overview}

This appendix is organized as follows.
Appendix~\ref{app:results} reports additional quantitative results: the attribution of grounding accuracy (\ref{app:grounding-acc}), teacher--student token agreement (\ref{app:agreement}), the multi-future protocol and the comparison with diffusion-based generation (\ref{app:diffusion}), privileged-cue variants (\ref{app:cues}), comparisons in multi-human, dynamic scenes (\ref{app:interactive}), robustness to segmentation noise (\ref{app:noise}), sensitivity to the intent-extraction VLM (\ref{app:vlm-sens}), runtime (\ref{app:runtime}), and soft token aggregation (\ref{app:soft-selector}).
Appendix~\ref{app:additional-qualitative} provides additional qualitative results.
Appendices~\ref{app:tokens}--\ref{app:impl} detail candidate-token construction, privileged supervision, the architecture, and training.
Appendix~\ref{app:broader} discusses broader impacts.

Table~\ref{tab:info_avail} summarizes which inputs each component of SGIP observes.
The VLM, the LLM, and the privileged teacher are used only to construct training-time supervision; only the student is deployed.

\begin{table}[h]
\centering
\small
\caption{Information availability across SGIP components. $c_i=(\bm{\phi}_i,\bm{e}_i,\bm{m}_i)$ denotes a candidate token.}
\label{tab:info_avail}
\setlength{\tabcolsep}{3pt}
\resizebox{\linewidth}{!}{%
\begin{tabular}{lll}
\toprule
\textbf{Component} & \textbf{Used at} & \textbf{Inputs} \\
\midrule
VLM intent extractor (offline) & training only & training video, 3D motion visualization, intent taxonomy \\
LLM rubric scorer (offline)    & training only & intent $\mathcal{I}$, candidate semantic labels, scene context \\
Privileged teacher             & training only & $M_{1:t}$, $S$, $\{c_i\}$, $\mathcal{I}$, cached $\{s^{\text{rubric}}_i\}$ \\
Observation-only student       & training + inference & $M_{1:t}$, $S$, $\{c_i\}$ \\
GT interaction target          & evaluation only & never input to any network \\
\bottomrule
\end{tabular}}
\end{table}

The teacher never observes the ground-truth grounding token; it observes only the candidate tokens, their semantic labels, the textual intent $\mathcal{I}$, and the geometric descriptors $\{\bm{\phi}_i\}$.
Distillation therefore transfers \emph{intent-aware reasoning} over candidate properties rather than ground-truth label leakage.

\FloatBarrier
\section{Additional Quantitative Results}
\label{app:results}

All SGIP results in this section are averaged over five random seeds unless stated otherwise. P/D denote path/destination.

\subsection{Attribution of Grounding Accuracy}
\label{app:grounding-acc}

\paragraph{Evaluation target.}
Grounding accuracy is measured against the GIMO-annotated interaction target, which is used only for evaluation (Table~\ref{tab:info_avail}).
We match each annotated target to OpenIns3D proposals by IoU on point membership (threshold $0.5$).
After cone filtering, the matched target remains in the candidate set for every evaluation sequence (Appendix~\ref{app:cone}).

\paragraph{Matched controls.}
To separate the effect of intent-conditioned supervision from that of the learnable head and candidate preprocessing (main Table~\ref{tab:grounding_acc}), all student variants use the same grounding MLP, input features, candidate sets, cone filtering, downstream architecture, and training protocol; only the prior-distillation target changes.
\emph{No prior KD} removes $\Lz$ and trains the same head with the remaining objectives; $\mathrm{KL}(R)$ and $\mathrm{KL}(G)$ distill the rubric-only and geometry-only priors (Eq.~\ref{eq:teacher_prior} with only one score), and $\mathrm{KL}(G{+}R)$ distills the fused teacher prior $\pi^{\Tch}$.
With the identical head, distilling $G$ yields 51\% Acc@1, whereas distilling $G{+}R$ yields 76\%; the No-prior-KD result (41\%) further shows that the head alone does not account for the accuracy of the deployed student.

\paragraph{Rank distribution.}
Table~\ref{tab:grounding_rank} reports, for the fixed priors and the deployed student, the fraction of sequences in which the annotated target is ranked at position $k$.
The unimodal priors reflect two complementary failure modes: the rubric score cannot disambiguate same-category instances (e.g., several chairs), while the geometric score is intent-blind and confuses objects that are equally reachable but not intent-compatible.
Fusing both concentrates mass on the target (86\% at rank 1), and the distilled student retains most of this concentration (76\%).

\begin{table}[h]
\centering
\small
\caption{Rank distribution of the GIMO-annotated target (\%). ``Rank $k$'' is the fraction of sequences in which the target is ranked exactly at position $k$; ``$\le 3$'' is their sum.}
\label{tab:grounding_rank}
\begin{tabular}{lcccc}
\toprule
Prior & Rank 1 & Rank 2 & Rank 3 & $\le 3$ \\
\midrule
Rubric only ($R$)              & 54 & 24 & 2  & 80 \\
Geometry only ($G$)            & 51 & 16 & 11 & 78 \\
Teacher $\pi^{\Tch}$ ($G{+}R$) & \textbf{86} & 5 & 3 & \textbf{94} \\
Student $\pi^{\Stu}$           & 76 & 5 & 3 & 84 \\
\bottomrule
\end{tabular}
\end{table}

\subsection{Teacher--Student Token Agreement}
\label{app:agreement}

To test the grounding-misalignment explanation of Section~\ref{sec:ablation}, we measure exact Top-1 token agreement $i^*_{\Tch}=i^*_{\Stu}$ between the frozen teacher and the student.
Agreement counts identical selections regardless of whether the selected token matches the ground truth.

\begin{table}[h]
\centering
\small
\caption{Teacher--student Top-1 token agreement, student grounding accuracy, and trajectory error on GIMO (mm).}
\label{tab:agreement}
\begin{tabular}{lccc}
\toprule
Student loss & T--S Agree@1 & GT Acc@1 & Traj.\ P/D $\downarrow$ \\
\midrule
$\LY$ only        & 43\% & 41\% & 856/896 \\
$\Lz$ only        & 70\% & 76\% & 574/688 \\
$\Lz + \LY$       & 70\% & 76\% & \textbf{549/681} \\
\bottomrule
\end{tabular}
\end{table}

Under $\LY$ alone, the teacher and student select different tokens in 57\% of the sequences, so the student imitates trajectories conditioned on a grounding it does not share.
Adding $\Lz$ raises agreement to 70\% and accuracy to 76\%.
Once grounding is aligned, adding $\LY$ further reduces trajectory error from 574/688 to 549/681\,mm without changing either grounding metric, i.e., motion-level distillation helps only after prior distillation has aligned the two branches.

\subsection{Multi-Future Protocol and Comparison with Diffusion-Based Generation}
\label{app:diffusion}

\paragraph{Metrics.}
Let $\{\hat{\bm{p}}^{(k)}_\tau\}_{k=1}^{K}$ be the predicted global root positions of $K$ futures and $\bm{p}_\tau$ the ground truth, $\tau=t{+}1,\dots,T$.
With $\mathrm{ADE}^{(k)}=\frac{1}{\Delta T}\sum_\tau\|\hat{\bm{p}}^{(k)}_\tau-\bm{p}_\tau\|_2$ and $\mathrm{FDE}^{(k)}=\|\hat{\bm{p}}^{(k)}_T-\bm{p}_T\|_2$,
\begin{gather}
\mathrm{minADE}@K=\min_k \mathrm{ADE}^{(k)},\qquad
\mathrm{minFDE}@K=\min_k \mathrm{FDE}^{(k)},\\
\mathrm{FDE@minADE}=\mathrm{FDE}^{(k^\star)},\quad k^\star=\arg\min_k \mathrm{ADE}^{(k)},
\end{gather}
\begin{equation}
\mathrm{APD}=\frac{2}{K(K-1)}\sum_{k<l}\frac{1}{\Delta T}\sum_{\tau}\bigl\|\hat{\bm{p}}^{(k)}_\tau-\hat{\bm{p}}^{(l)}_\tau\bigr\|_2 .
\end{equation}
FDE@minADE evaluates whether the trajectory that is best on average also arrives at the right place, which minFDE alone does not guarantee.

\paragraph{SGIP futures.}
SGIP decodes each of the top-$K$ tokens of $\pi^{\Stu}$ separately; the observed motion, scene, and model parameters are fixed and only the selected token's mask and embedding change.
SGIP therefore produces $K$ futures without privileged cues or stochastic sampling, each associated with an interpretable scene token.
Main Table~\ref{tab:multifuture} reports SGIP for $K\in\{1,2,3\}$.
Diffusion-based generation models offer an alternative route to multiple futures, obtaining them by stochastic sampling rather than from an explicit distribution over scene tokens; below, we compare SGIP with such a model under the same inputs and the same prediction budget.

\paragraph{AffordMotion adaptation.}
AffordMotion~\cite{wang2024affordmotion} is a two-stage conditional diffusion model that first predicts a scene affordance map and then generates motion conditioned on it.
Since it is designed for language-guided motion generation rather than forecasting, we adapt it to HMP following GAP3DS~\cite{yu2025vision} by replacing the language condition with an encoding of the observed motion, so that it receives the same inputs ($M+S$) as SGIP; both methods use the same GIMO split, windows, and evaluation code.
For single-future evaluation, AffordMotion produces one diffusion sample per sequence and SGIP decodes its Top-1 token; for $K=3$, AffordMotion draws three independent samples and SGIP decodes its Top-3 tokens.

\begin{table}[h]
\centering
\small
\caption{Comparison with the diffusion-based AffordMotion adapted to HMP. Both methods use $M+S$ at inference. Single-future errors are in mm; multi-future root-trajectory errors are in m.}
\label{tab:diffusion_full}
\setlength{\tabcolsep}{4pt}
\begin{tabular}{lccccc}
\toprule
& \multicolumn{2}{c}{Single future} & \multicolumn{3}{c}{Multi-future ($K=3$)} \\
\cmidrule(lr){2-3}\cmidrule(lr){4-6}
Method & Traj.\ P/D $\downarrow$ & MPJPE P/D $\downarrow$ & minADE $\downarrow$ & minFDE $\downarrow$ & FDE@minADE $\downarrow$ \\
\midrule
AffordMotion & 577/714 & 160.9/\textbf{194.1} & 0.597 & 0.728 & 0.730 \\
SGIP         & \textbf{549/681} & \textbf{153.1}/195.1 & \textbf{0.467} & \textbf{0.523} & \textbf{0.571} \\
\bottomrule
\end{tabular}
\end{table}

Under the matched setting, SGIP achieves lower single-future trajectory errors and MPJPE-path, and with three predictions it attains lower minADE, minFDE, and FDE@minADE than three diffusion samples.
Rather than relying on stochastic sampling, SGIP exposes an explicit distribution over interpretable scene tokens, provides a deterministic Top-1 prediction, and generates controllable alternatives by changing the selected token.

\subsection{Privileged-Cue Variants}
\label{app:cues}

\paragraph{Cue interface.}
All privileged variants inject their cue only through scores over the same candidate tokens, followed by the same Top-1 selection and the same token-conditioned predictor; no cue is fed to the motion predictor directly.
For intent, $s^{\mathcal{I}}_i = s^{\text{rubric}}_i+\alpha s^{\text{geo}}_i$ (Eq.~\ref{eq:teacher_prior}).
For gaze, we use the single 3D gaze point at the final observed frame and compute a proximity score $s^{\text{gaze-dist}}_i$ between candidate $c_i$ and the gaze point and a directional-alignment score $s^{\text{gaze-align}}_i$ between the head-to-gaze and head-to-candidate directions:
\begin{align}
s^{g}_i &= \alpha s^{\text{geo}}_i + \lambda_d\, s^{\text{gaze-dist}}_i + \lambda_a\, s^{\text{gaze-align}}_i, \\
s^{g+\mathcal{I}}_i &= s^{\mathcal{I}}_i + \lambda_d\, s^{\text{gaze-dist}}_i + \lambda_a\, s^{\text{gaze-align}}_i,
\end{align}
with $\lambda_d=\lambda_a=3$ selected on the training split. The scores are normalized before selection.

\begin{table}[h]
\centering
\small
\caption{Privileged-cue variants on GIMO (mm). All variants share the same token-conditioned predictor.}
\label{tab:cue_variants}
\begin{tabular}{llcc}
\toprule
Variant & Cue & Traj.\ P/D $\downarrow$ & MPJPE P/D $\downarrow$ \\
\midrule
SGIP (deployed) & --  & 549/681 & 153.1/195.1 \\
$\textit{SGIP}^{\dagger}_{\mathrm{priv}}$ & gaze & 528/635 & 149.9/190.6 \\
$\textit{SGIP}^{\dagger}_{\mathrm{priv}}$ & intent & 485/556 & 151.2/187.8 \\
$\textit{SGIP}^{\dagger}_{\mathrm{priv}}$ & gaze+intent & \textbf{475/539} & \textbf{149.7/180.8} \\
\bottomrule
\end{tabular}
\end{table}

Intent yields lower trajectory error than gaze, and combining both cues improves all four metrics over either cue alone, consistent with intent providing semantic evidence about \emph{which} target will be engaged and gaze providing spatial evidence for localizing it.
Because the cues act only through the candidate scores, heterogeneous privileged cues can be combined without modifying the downstream predictor.

\subsection{Comparison in Multi-Human, Dynamic Scenes}
\label{app:interactive}

GIMO and PROX contain a single person per scene.
To examine SGIP in scenes where other people move around the target, we compare it with HUMOF~\cite{sun2026humof}, a forecasting model designed for interactive social scenes, on HIK~\cite{tanke2023hik} and HOI-M$^3$~\cite{zhang2024hoim3}.

\paragraph{Protocol.}
We follow the target-centric forecasting setting of HUMOF~\cite{sun2026humof}, in which each sample predicts the future motion of one target person in a multi-person scene.
For HIK~\cite{tanke2023hik}, recordings A--C are used for training and D for testing, with 25 observed and 50 future frames at 25\,FPS.
For HOI-M$^3$~\cite{zhang2024hoim3}, we use a subset of 9/3 train/test sequences (3{,}332/689 samples) with 30 observed and 60 future frames at 30\,FPS.
We train HUMOF with its official implementation on the same samples as SGIP, and evaluate both with the released Path/Pose metrics, using the last observed root of the target as the local origin.

\paragraph{Inputs.}
HUMOF observes the temporal histories of neighboring humans ($H$) in addition to the target history $M$ and the scene $S$.
SGIP uses only the target history and represents neighboring humans and objects as static tokens at the last observed frame.
Object tokens on HIK and HOI-M$^3$ are constructed from native object annotations and geometry at the last observed frame. 
HIK uses native action annotations as training-time intents for a Qwen3-14B text teacher. For HOI-M$^3$, Qwen3-VL-8B infers intents from a video and a reference image identifying the target person; future frames provide privileged teacher supervision.

\begin{table}[h]
\centering
\scriptsize
\caption{Comparison in multi-human, dynamic scenes on HIK and HOI-M$^3$ (mm). Both methods are trained and evaluated on identical samples.}
\label{tab:interactive_full}
\setlength{\tabcolsep}{2.6pt}
\resizebox{\linewidth}{!}{%
\begin{tabular}{llc|ccccc|ccccc}
\toprule
& & & \multicolumn{5}{c|}{Path error} & \multicolumn{5}{c}{Pose error} \\
Dataset & Method & $H$ & 0.5s & 1.0s & 1.5s & 2.0s & mean & 0.5s & 1.0s & 1.5s & 2.0s & mean \\
\midrule
\multirow{2}{*}{HIK}
& HUMOF~\cite{sun2026humof} & \cmark & 96.9 & 231.3 & 400.3 & 546.6 & 251.9 & \textbf{73.5} & \textbf{99.3} & \textbf{116.7} & \textbf{123.7} & \textbf{90.6} \\
& SGIP & \xmark & \textbf{89.2} & \textbf{130.9} & \textbf{167.7} & \textbf{205.7} & \textbf{127.1} & 81.5 & 114.9 & 130.7 & 138.8 & 103.1 \\
\midrule
\multirow{2}{*}{HOI-M$^3$}
& HUMOF~\cite{sun2026humof} & \cmark & \textbf{104.7} & \textbf{238.0} & \textbf{385.8} & \textbf{538.4} & \textbf{253.9} & \textbf{117.2} & 163.0 & 189.3 & 210.7 & 148.9 \\
& SGIP & \xmark & 134.9 & 280.7 & 437.4 & 583.0 & 291.1 & 121.1 & \textbf{159.2} & \textbf{179.1} & \textbf{200.9} & \textbf{145.0} \\
\bottomrule
\end{tabular}}
\end{table}

\paragraph{Discussion.}
On HIK, SGIP achieves lower path error than HUMOF at every horizon, reducing the mean from 251.9 to 127.1\,mm without observing neighboring-human histories, while HUMOF, which explicitly models these histories, attains lower pose error.
On HOI-M$^3$, SGIP obtains lower pose error from 1.0\,s onward and on average.
These results indicate that the scene-grounded token representation transfers to multi-human, dynamic scenes even without neighbor histories; modeling time-varying human and object states is a natural extension.

\subsection{Robustness to Segmentation Noise}
\label{app:noise}

\paragraph{Protocol.}
We corrupt the OpenIns3D output at test time on all GIMO evaluation sequences while keeping the trained model fixed.
We perturb instance-mask logits with Gaussian noise $\sigma\in\{0.5,1.0,2.0\}$ and simulate under- and over-segmentation by merging the target with its nearest object and splitting it into two contiguous instances, respectively.
Gaussian results are averaged over three corruption seeds; merge and split are deterministic.
\emph{Target recall} indicates whether a target-derived candidate survives corruption and cone filtering.
\emph{Target Acc@1} is computed over all evaluation sequences, counting missing targets as incorrect; for merge/split, the merged candidate or either fragment counts as target-derived.

\begin{table}[h]
\centering
\scriptsize
\caption{Segmentation-noise ablation on GIMO. Gaussian rows report mean$\pm$std over three corruption seeds; motion errors are in mm.}
\label{tab:noise}
\setlength{\tabcolsep}{4.5pt}
\begin{tabular}{lcccc}
\toprule
Condition & Target recall (\%) & Target Acc@1 (\%) $\uparrow$ & Traj.\ P/D $\downarrow$ & MPJPE P/D $\downarrow$ \\
\midrule
Clean & 100.00 & 75.68 & 549.0/681.0 & 153.1/195.1 \\
\midrule
Gaussian, $\sigma=0.5$ & $100.00\pm0.00$ & $69.37\pm1.56$ & $559.0\pm14.7$/$708.0\pm17.4$ & $158.1\pm1.1$/$211.1\pm1.2$ \\
Gaussian, $\sigma=1.0$ & $99.10\pm1.56$  & $65.77\pm1.56$ & $545.0\pm15.0$/$690.0\pm35.5$ & $158.5\pm0.9$/$212.1\pm1.5$ \\
Gaussian, $\sigma=2.0$ & $81.08\pm4.68$  & $53.15\pm1.56$ & $581.0\pm20.3$/$749.0\pm34.1$ & $158.1\pm1.6$/$210.3\pm0.4$ \\
Target--neighbor merge & 100.00 & 56.76 & 661.0/875.0 & 159.2/211.3 \\
Target two-way split   & 100.00 & 54.05 & 644.0/908.0 & 157.1/208.8 \\
\bottomrule
\end{tabular}
\end{table}

Under mild noise ($\sigma=0.5$), target recall remains 100\% while Target Acc@1 decreases from 75.68\% to 69.37\% and trajectory error increases from 549/681 to 559/708\,mm.
Under severe noise ($\sigma=2.0$), target recall and Target Acc@1 drop to 81.08\% and 53.15\%.
Merge and split preserve a target-derived candidate but raise destination error to 875 and 908\,mm, since the merged or fragmented instance no longer delineates the interaction surface.
Severe and structural segmentation errors therefore cause substantially larger degradation than mild mask corruption, which motivates the limitation discussed in Section~\ref{sec:conclusion}.

\subsection{Sensitivity to the Intent-Extraction VLM}
\label{app:vlm-sens}

The main experiments use Qwen3-VL-8B~\cite{bai2025qwen3vl} to extract sequence-level intents for both GIMO and PROX.
We replace it with Qwen3-VL-4B and Idefics3-8B~\cite{laurencon2024idefics3}, regenerate the intent labels, and retrain the same teacher--student pipeline while keeping the dataset-specific taxonomy, candidate tokens, architecture, optimization, and evaluation protocol fixed.

\begin{table}[h]
\centering
\small
\caption{Sensitivity to the VLM used for training-time intent. Agreement is the exact intent-category agreement with Qwen3-VL-8B within each dataset (mm).}
\label{tab:vlm_sens}
\setlength{\tabcolsep}{3.5pt}
\resizebox{\linewidth}{!}{%
\begin{tabular}{lccc|ccc}
\toprule
& \multicolumn{3}{c|}{GIMO} & \multicolumn{3}{c}{PROX} \\
\cmidrule(lr){2-4}\cmidrule(lr){5-7}
VLM & Agreement & Traj.\ P/D & MPJPE P/D & Agreement & Traj.\ P/D & MPJPE P/D \\
\midrule
Qwen3-VL-8B (default) & -- & 549/681 & 153.1/195.1 & -- & 470/576 & 194.6/197.7 \\
Qwen3-VL-4B & 90.7\% & 527/753 & 152.8/196.8 & 96.6\% & 459/570 & 195.0/209.5 \\
Idefics3-8B & 92.1\% & 584/723 & 154.9/190.6 & 97.0\% & 493/564 & 181.1/202.9 \\
\bottomrule
\end{tabular}}
\end{table}

The alternative extractors agree with Qwen3-VL-8B on 90.7--92.1\% of GIMO and 96.6--97.0\% of PROX sequences.
No extractor consistently improves or degrades all metrics across both datasets, indicating reasonable robustness to VLM scale and family, with larger variation in destination errors.

\subsection{Runtime}
\label{app:runtime}

We measure one-time scene preprocessing and per-observation student inference separately on a single NVIDIA RTX~3090 with batch size 1 and FP32.
Scene initialization, including point-cloud loading, OpenIns3D instance segmentation, and construction of the static instance inventory, takes $61.73\pm7.18$\,s per scene across the 19 GIMO scenes; the resulting masks and inventory are cached.

\begin{table}[h]
\centering
\small
\caption{Runtime on one RTX~3090 (batch size 1, FP32).}
\label{tab:runtime}
\begin{tabular}{lcc}
\toprule
Stage & Frequency & Runtime \\
\midrule
\multicolumn{3}{l}{\emph{One-time preprocessing}} \\
OpenIns3D scene initialization & once per scene & $61.73\pm7.18$\,s \\
\midrule
\multicolumn{3}{l}{\emph{Online student inference}} \\
Motion-dependent descriptors      & per observation & $59.72\pm18.99$\,ms \\
Student prior + Top-1 selection   & per observation & $6.10\pm0.40$\,ms \\
Host-to-device transfer           & per observation & $0.38\pm0.07$\,ms \\
Student motion prediction         & per observation & $66.54\pm1.04$\,ms \\
\midrule
Total online inference            & per observation & 132.75\,ms (7.53\,FPS) \\
\bottomrule
\end{tabular}
\end{table}

After scene initialization, SGIP requires 132.75\,ms per observation, below the 500\,ms observation interval of the 2-FPS GIMO protocol, so it supports online prediction in a static scene; processing a new scene incurs the initialization cost again.
No VLM or LLM is invoked at inference.

\subsection{Hard vs.\ Soft Grounding Token Aggregation}
\label{app:soft-selector}
\begin{figure}[t]
\centering
\includegraphics[width=0.85\textwidth]{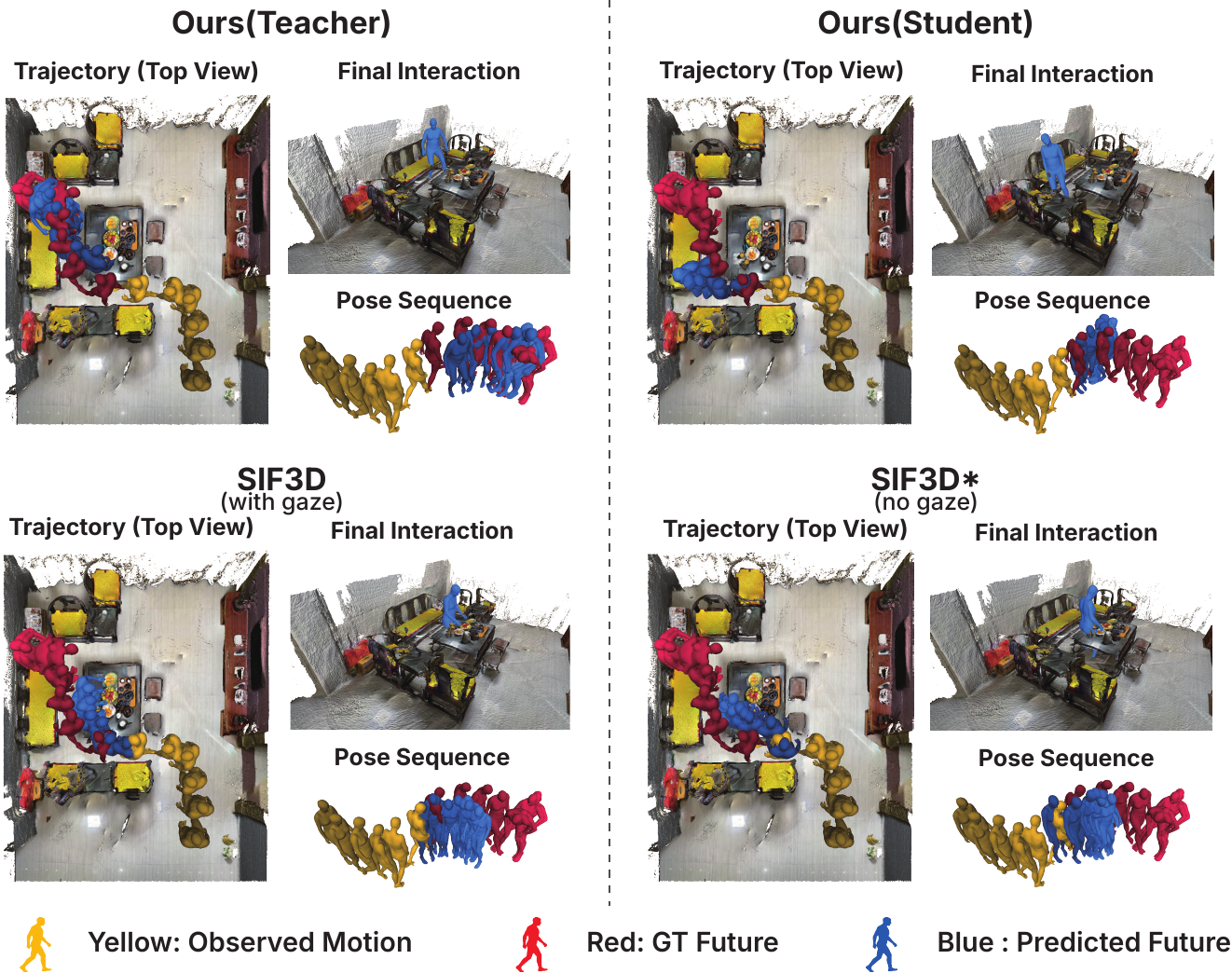}
\caption{\textbf{Qualitative comparison across SGIP branches and SIF3D.} For one cluttered GIMO test scene, given the same observed motion: \textbf{top left}, the privileged SGIP teacher ($\textit{SGIP}^\dagger_{\text{priv}}$, textual intent); \textbf{top right}, the deployed observation-only SGIP student; \textbf{bottom left}, SIF3D~\citep{lou2024sif3d} with gaze; \textbf{bottom right}, SIF3D$^{*}$ without gaze. Each panel shows the top-down trajectory, the final interaction frame, and the pose sequence. Yellow: observed motion; red: GT future; blue: predicted future.}
\label{fig:qual-comparison}
\end{figure}

The motion predictor consumes a single grounding token $i^{*}_{b}$ via its mask $\bm{m}_{i^{*}_{b}}$ and embedding $\bm{e}_{i^{*}_{b}}$ (Eqs.~\ref{eq:traj}--\ref{eq:pose_branch}).
A natural alternative replaces the hard argmax (Eq.~\ref{eq:select}) with a probability-weighted mixture of the top-$k$ tokens:
\begin{equation}
\bar{\bm{m}} = \sum_{i \in \mathcal{K}_k} \tilde{\pi}^{\Stu}_{i}\,\bm{m}_{i},
\qquad
\bar{\bm{e}} = \sum_{i \in \mathcal{K}_k} \tilde{\pi}^{\Stu}_{i}\,\bm{e}_{i},
\label{eq:soft-aggregation}
\end{equation}
where $\mathcal{K}_k = \mathrm{TopK}_k(\pi^{\Stu})$ and $\tilde{\pi}^{\Stu}_{i} = \pi^{\Stu}_{i} / \sum_{j \in \mathcal{K}_k} \pi^{\Stu}_{j}$; $D^{\Stu}$ then consumes $(\bar{\bm{m}}, \bar{\bm{e}})$ in place of $(\bm{m}_{i^{*}_{\Stu}}, \bm{e}_{i^{*}_{\Stu}})$.
The soft mask $\bar{\bm{m}}$ enters the spatial importance map in place of $\bm{m}_{i^*_{\Stu}}$ (Eq.~\ref{eq:spatial_importance}).
All variants share the same trained student; only the token selection differs at inference.

Soft aggregation smooths the conditioning signal under ambiguity but dilutes the geometry of the selected region by blending in less likely tokens.
When the blended tokens lie in different parts of the scene, $\bar{\bm{e}}$ no longer corresponds to any single target and $\bar{\bm{m}}$ covers several disjoint regions, so the predicted trajectory tends toward an average of the candidate destinations and the arrival pose is not anchored to one contact surface.
Accordingly, the trajectory degradation in main Table~\ref{tab:hard_vs_soft} grows as more tokens are blended ($k=2\to3$).
When multiple futures are desired, decoding the top-$k$ tokens separately (Section~\ref{sec:multifuture}) preserves the multimodal structure of $\pi^{\Stu}$ without blending.

\FloatBarrier
\section{Additional Qualitative Results}
\label{app:additional-qualitative}

\subsection{Teacher, Student, and Gaze-Based Baselines}
\label{app:qual-comparison}
Figure~\ref{fig:qual-comparison} compares the two SGIP branches with SIF3D on the same scene.
Without gaze, SIF3D drifts toward an incorrect region and produces an inconsistent arrival pose; with gaze, it recovers the rough direction but still over- or under-shoots the target, indicating that gaze alone does not fully disambiguate the interaction target in cluttered scenes.
Both SGIP branches ground the trajectory on the correct target and produce a coherent arrival pose, and the observation-only student closely tracks the privileged teacher despite using neither gaze nor textual intent at inference.

\subsection{Intent-Consistent Alternative Interactions}
\label{app:alternative}

\begin{figure}[t]
\centering
\includegraphics[width=\textwidth]{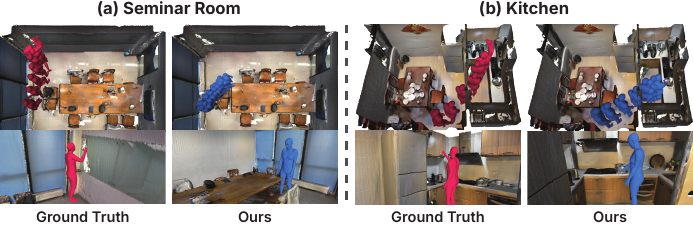}
\caption{\textbf{Intent-consistent alternative interactions on GIMO.} For each scene, the recorded future (red) and the SGIP prediction (blue) are shown as top-down trajectories (top) and final-frame poses (bottom). \textbf{(a)}~Seminar room: the recorded person goes to retrieve an eraser from the blackboard, while SGIP selects a desk as its grounding target. \textbf{(b)}~Kitchen: the ground-truth target is the cabinet, while SGIP selects a pot on the dining table. These examples illustrate intent-aware grounding of plausible alternative futures beyond the recorded interaction. In both cases, SGIP commits to a different but reachable target and completes a coherent interaction with it.}
\label{fig:alternative}
\end{figure}

Single-future metrics count every deviation from the recorded motion as an error, yet in cluttered scenes the same observed motion often leaves several intentions open.
Figure~\ref{fig:alternative} shows cases in which the student's top-ranked token differs from the recorded target.
Rather than drifting between candidates, stopping short, or averaging several destinations, SGIP commits to an alternative object that is reachable from the observed approach and that affords a common everyday activity, and it completes the interaction with that object: the trajectory ends at the selected object and the arrival pose is oriented toward it.
Such predictions are consistent with an intent that the observation alone does not rule out, and they arise mainly when several targets are similarly compatible with the observed approach and the observation window ends before the person commits to one of them.
This behavior follows from making grounding explicit: each prediction corresponds to one interpretable target together with a physically coherent interaction, which is also why SGIP exposes ranked alternatives (Section~\ref{sec:multifuture}) instead of a single averaged future.

\FloatBarrier
\section{Candidate Grounding Tokens}
\label{app:tokens}

This appendix details the construction of the candidate grounding token set $\mathcal{C} = \mathcal{C}^{\text{obj}} \cup \mathcal{C}^{\text{nav}}$ used by the prior (Section~\ref{sec:problem}).

\subsection{Object-Instance Tokens}
\label{app:tokens-obj}

\begin{figure}[t]
\centering
\includegraphics[width=0.8\textwidth]{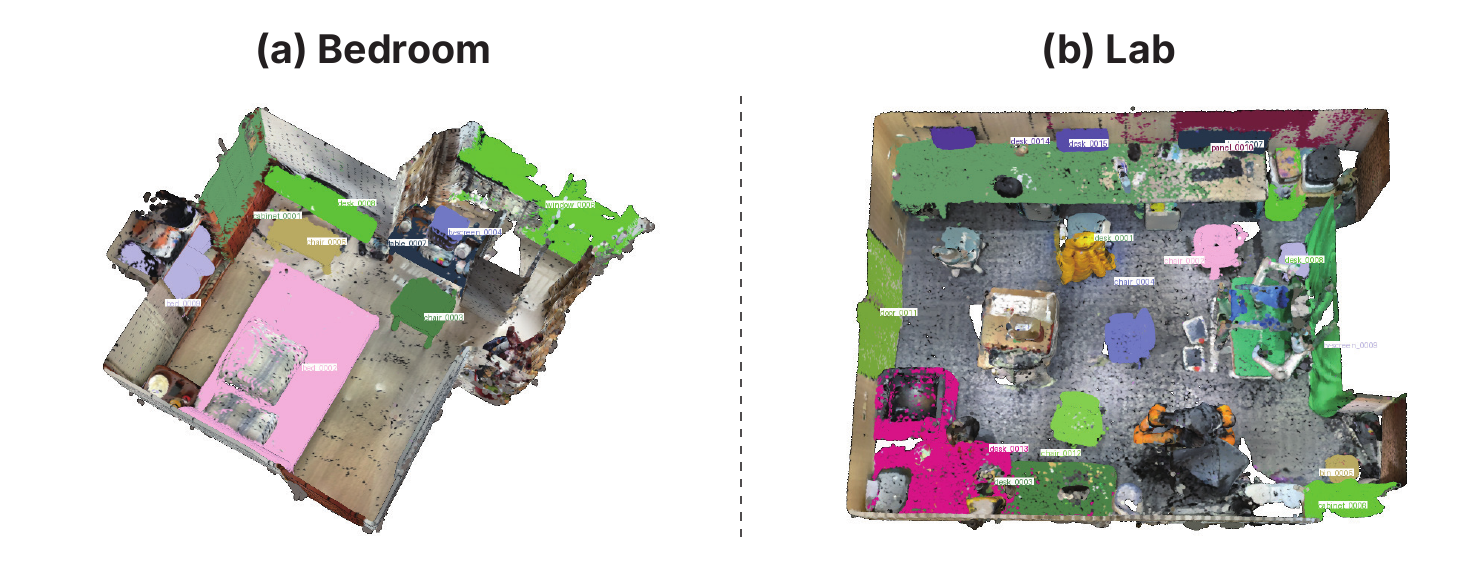}
\caption{\textbf{Zero-shot 3D instance segmentation by OpenIns3D~\cite{huang2024openins3d}} on two GIMO scenes. Each color denotes a distinct object instance $c_i \in \mathcal{C}^{\text{obj}}$; same-category objects (e.g., multiple chairs) are separated into distinct instances.}
\label{fig:openins3d}
\end{figure}

Object instances are obtained by zero-shot 3D instance segmentation on the scene point cloud $S$ with OpenIns3D~\cite{huang2024openins3d}, producing per-instance point-cloud masks $\bm{m}_i \subseteq S$ with semantic labels (Figure~\ref{fig:openins3d}).
Same-category objects (e.g., two chairs) share a label but remain distinct tokens with distinct masks; the geometric compatibility score (Appendix~\ref{app:geo_score}) disambiguates among them.
The unfiltered inventory contains $14.68\pm6.01$ object tokens per GIMO evaluation scene.

\subsection{Forward-Facing Cone Filtering}
\label{app:cone}

The cone is an SGIP-specific candidate-pruning step computed only from the observed motion; it uses neither gaze nor target annotations.
At each observed frame $\tau$, we define a cone $C(x_\tau)$ originating at the head whose axis bisects the head- and torso-facing directions, with a fixed aperture of $120^\circ$, and take the union over the observation window:
\begin{equation}
C(M_{1:t}) = \bigcup_{\tau=1}^{t} C(x_\tau).
\end{equation}
An object token is retained if its mask intersects the aggregated cone, i.e., $\bm{m}_i \cap C(M_{1:t}) \neq \emptyset$.
Since upcoming interactions are typically preceded by head and body alignment toward the target~\cite{land2001ways, hayhoe2005eye, zheng2022gimo}, the $120^\circ$ cone approximates a broad forward-facing interaction region and removes rearward objects that are unlikely to be immediate interaction targets.
The teacher, the student, and all SGIP ablations use the same filtered candidate set.

Because the filtered set depends on the observed motion, the number of retained candidates varies across sequences.
We first average sequence-level counts within each scene and then summarize them across the 19 GIMO evaluation scenes (Table~\ref{tab:cone}).
Together with the always-retained navigation token (Appendix~\ref{app:tokens-nav}), this yields $13.60\pm4.62$ grounding candidates on average, and the annotated target remains in the filtered set for every evaluation sequence (100\% target recall).

\begin{table}[h]
\centering
\small
\caption{Cone-filtered object candidates on GIMO. Sequence-level counts are first averaged within each scene; values are mean$\pm$std across the 19 evaluation scenes.}
\label{tab:cone}
\begin{tabular}{lcc}
\toprule
Candidate set & Object tokens & Retention (\%) \\
\midrule
Before cone filtering & $14.68\pm6.01$ & 100.00 \\
After cone filtering ($120^\circ$) & $12.60\pm4.62$ & $88.83\pm11.07$ \\
\bottomrule
\end{tabular}
\end{table}

\subsection{Navigation-Region Token}
\label{app:tokens-nav}

The navigation token $\mathcal{C}^{\text{nav}}$ represents object-agnostic free space (e.g., open floor, doorways, corridors).
Unlike object tokens, it is scene-global and is not cone-filtered, so that the prior can represent free-space motion when the observed motion is not yet aligned with any specific object.
Its mask is the complement of all object masks within an interaction-relevant region.

\subsection{Geometric Descriptor}
\label{app:arch:descriptor}

The geometric descriptor concatenates two groups of features,
$\bm{\phi}_i = [\bm{\phi}_{\mathrm{geo}}(c_i),\ \bm{\phi}_{\mathrm{rel}}(c_i, M_{1:t})] \in \mathbb{R}^F$ with $F = 16$.
$\bm{\phi}_{\mathrm{geo}}(c_i)$ encodes token-intrinsic spatial properties (centroid, oriented bounding-box size, height above the floor, and vertical orientation of the contact normal).
$\bm{\phi}_{\mathrm{rel}}(c_i, M_{1:t})$ encodes motion-relative quantities at the last observed frame: the signed distance from the root to $c_i$, the root-to-token approach angle, root-horizontal alignment, face-normal alignment, and torso alignment.
Features are z-score normalized with training-set statistics.

\subsection{Geometric Compatibility Score}
\label{app:geo_score}

Each alignment term in Eq.~\eqref{eq:geo} measures the cosine similarity between a motion-derived direction at the last observed frame and the direction from the root toward token $c_i$, normalized to $[0, 1]$:
$\mathrm{align}_{\text{face}}$ uses the head-facing direction, $\mathrm{align}_{\text{body}}$ the torso-facing direction, and $\mathrm{align}_{\text{line}}$ the recent trajectory direction.
We fix $w_{\text{face}} = 0.5$, $w_{\text{body}} = 2$, and $w_{\text{line}} = 0.5$ across all experiments, selected on a held-out portion of the training split.

\FloatBarrier
\section{Privileged Supervision}
\label{app:supervision}

\subsection{VLM-Based Intent Extraction}
\label{sec:appendix-intent-extraction}

We label the intent of each training sequence with a frozen Qwen3-VL-8B~\cite{bai2025qwen3vl}.
Because a person's intent is difficult to infer from a single viewpoint, the VLM jointly receives two videos of the full training clip.
On GIMO, these are the \textbf{egocentric video}, which conveys what the person looks at and immediate interaction targets, and a \textbf{rendering of the person's motion in the reconstructed 3D scene}, which conveys where the action takes place.
On PROX, which has no egocentric video, the first input is the RGB video recorded by the static PROX camera.
The prompt includes (i) the definitions of the dataset's intent categories, (ii) the role of each video, and (iii) an instruction to answer \textit{unknown} when no category applies.
The predicted category is converted into a textual intent $\mathcal{I}$ (e.g., ``take some rest.'') that is used only by the privileged teacher.
The VLM, the videos, and the pseudo-intents are used only to construct training-time supervision.

\subsection{Intent Taxonomies}
\label{sec:intent-taxonomy}

\paragraph{GIMO (16 categories).}
\textit{sit}, \textit{sleep}, \textit{drink}, \textit{eat}, \textit{read}, \textit{use}, \textit{take}, \textit{open}, \textit{light}, \textit{clean}, \textit{dress}, \textit{throw}, \textit{cook}, \textit{write}, \textit{fix}, and \textit{unknown}.

\paragraph{PROX (7 categories).}
{\textit{sit-up}, \textit{sit-down}, \textit{lie}, \textit{walk}, \textit{touch}, \textit{jump}, and \textit{unknown}.}

\paragraph{Design principles.}
Categories are intended to be mutually exclusive while covering the main behavioral patterns of each dataset, and each corresponds to a unit action that is visually distinguishable from the available views, so that the VLM can classify it from visual cues alone.
The explicit \textit{unknown} class prevents ambiguous sequences from being forced into an incorrect category.
Robustness to the choice of VLM is reported in Appendix~\ref{app:vlm-sens}.

\subsection{LLM-Based Rubric Scorer}
\label{app:llm}

\paragraph{Rubric.}
A reasoning LLM scores each candidate token $c_i$ on two integer criteria:
\begin{align}
a_i &\in \{1, \dots, 5\} &&\text{(\emph{intent affordance}: how well does $c_i$ semantically support $\mathcal{I}$?)} \nonumber\\
u_i &\in \{1, \dots, 5\} &&\text{(\emph{intent usability}: given the scene, how easily can $\mathcal{I}$ be realized using $c_i$?)} \nonumber
\end{align}
The total $s^{\mathrm{rubric}}_i = a_i + u_i \in [2, 10]$ enters the temperature softmax of Eq.~\eqref{eq:teacher_prior}.
Geometric alignment is already available to the student through $\bm{\phi}_i$ and $\bm{h}_M$, so the rubric deliberately focuses on intent-aware semantic judgment, which is the teacher's distinctive contribution.

\paragraph{LLM call.}
We use Qwen3-14B~\cite{yang2025qwen3} with NF4 4-bit quantization and greedy decoding (temperature $0$, top-$p=1$, at most 4096 new tokens, fixed seed).
The prompt enforces a two-stage output: a natural-language rationale (used to encourage scoring consistency, not parsed as supervision) followed by a strict JSON block listing per-token integer scores and the recomputed total.
Each (sequence, candidate set) pair is queried once and cached offline; no LLM call occurs during training updates or inference.

\paragraph{JSON validation and exclusion.}
Each JSON block is validated against a schema that checks per-criterion ranges and recomputes $s^{\mathrm{rubric}}_i = a_i + u_i$ when the reported total disagrees.
Out-of-range scores, missing entries, or duplicate token identifiers trigger a single retry; duplicated score values or repeated object names are legitimate (e.g., two chairs) and do not.
The protocol was designed to exclude the entire sample in the event of a JSON parsing failure, since removing individual tokens could alter the candidate set and eliminate the target. In practice, no parsing failures occurred, and no training samples were excluded ($N=0$).



\FloatBarrier
\section{Architecture Details}
\label{app:arch}

\subsection{Encoders and Student Scoring Head}
\label{app:arch:encoders}

\paragraph{Motion encoder $E_M^b$.}
A 4-layer Transformer with hidden dimension $256$ and $4$ heads ingests the per-frame SMPL-X state of $M_{1:t}$ and outputs the motion feature $\bm{h}_M^b$.

\paragraph{Token encoder $E_{\mathcal{C}}$.}
For each token $c_i$, a shared PointNet++~\cite{qi2017pointnetpp} backbone (2 set-abstraction layers) extracts a point-cloud feature from the points of $\bm{m}_i$, which is concatenated with a semantic-label embedding and a projection of $\bm{\phi}_{\mathrm{geo}}(c_i)$ and passed through a two-layer MLP (GELU) to produce $\bm{e}_i \in \mathbb{R}^{d_e}$, $d_e = 16$.

\paragraph{Student scoring head $F^{\mathrm{grd}}_{\Stu}$.}
A permutation-equivariant set encoder over candidates. Given $\bm{h}_M^{\Stu}$ and $\{\bm{\phi}_i\}$:
\begin{enumerate}
  \item Per-token MLP: $\bm{u}_i = \mathrm{MLP}_{\mathrm{cand}}([\bm{\phi}_i; \bm{h}_M^{\Stu}]) \in \mathbb{R}^{d_h}$, $d_h = 256$, GELU.
  \item Masked mean and max pooling over candidates: $\bar{\bm{u}}^{\mathrm{mean}}, \bar{\bm{u}}^{\mathrm{max}} \in \mathbb{R}^{d_h}$.
  \item Context projection: $\bar{\bm{c}} = \mathrm{MLP}_{\mathrm{ctx}}([\bar{\bm{u}}^{\mathrm{mean}}; \bar{\bm{u}}^{\mathrm{max}}])$.
  \item Per-token logit: $F^{\mathrm{grd}}_{\Stu}(\bm{h}_M^{\Stu}, \bm{\phi}_i) = \mathrm{MLP}_{\mathrm{head}}([\bm{u}_i; \bar{\bm{c}}]) \in \mathbb{R}$.
\end{enumerate}
The logits are normalized with a masked softmax over valid candidates (Eq.~\ref{eq:student_softmax}).

\subsection{Motion Predictor}
\label{app:arch:decoder}

The motion predictor $D^b$ follows the trajectory-then-pose design of Section~\ref{sec:decode}; the student and the teacher use the same architecture and differ only in the source of the selected token.

\paragraph{Scene encoding.}
The scene point cloud $S$ is encoded by PointNet++ into per-point features $F_S \in \mathbb{R}^{N \times d}$.

\paragraph{Trajectory predictor.}
Each block applies self-attention to the motion tokens $H_M$, cross-attention to the scene features, and an MLP fusion with the selected object's time-indexed latent sequence:
\begin{align}
H' &= \mathrm{SelfAttn}(H_M), \\
C^{\text{TP}} &= \mathrm{softmax}\bigl(Q' K_S^\top / \sqrt{d}\bigr) V_S, \\
H^{\text{TP}} &= \mathrm{MLP}\bigl([H', C^{\text{TP}},
\bm{E}^{\text{obj}}_{i^*_b}]\bigr),
\end{align}
where $Q' = W_Q H'$, $K_S = W_K F_S$, and $V_S = W_V F_S$.
Here, $\bm{E}^{\text{obj}}_{i^*_b}$ is produced by encoding the selected object's point cloud into a sequence of time-indexed latents. At each temporal position, the corresponding object latent is concatenated with the attention features before MLP fusion. The same object latent sequence is supplied to each trajectory block.
A linear head maps the output of the last block to the future root trajectory $\hat{\bm{r}}^b_{t+1:T}$ (translation and orientation).

\paragraph{Spatial importance map.}
The predicted trajectory induces a spatial importance map over scene points (Eq.~\ref{eq:spatial}):
\begin{equation}
P^b = f_{\text{dist}}(S, \hat{\bm{r}}^b)
+ f_{\text{dir}}(S, \hat{\bm{r}}^b)
+ f_{\text{mlp}}(S, \hat{\bm{r}}^b)
+ \delta_{\text{obj}}\,
f_{\text{obj}}(S,\bm{c}_{i^*_b}, \hat{\bm{r}}^b),
\label{eq:spatial_importance}
\end{equation}
where $f_{\text{dist}}$ is a path-proximity prior, $f_{\text{dir}}$ a forward-facing alignment prior, $f_{\text{mlp}}$ a learned spatial bias, and
$f_{\text{obj}}$ an object-direction alignment prior scaled by a learned coefficient. It computes the horizontal cosine similarity between the root-to-scene-point and root-to-object-center directions, where $\bm{c}_{i^*_b}$ is the centroid of the selected object's point cloud. The gate $\delta_{\text{obj}} \in \{0,1\}$ equals $1$ when a valid point cloud for the selected object is available and $0$ otherwise. This gate affects only the object-direction term; the remaining spatial-prior terms are retained.

\paragraph{Pose predictor.}
The pose branch first injects the predicted trajectory into the motion tokens and then performs scene cross-attention with $P^b \in \mathbb{R}^{\Delta T \times N}$ as an additive bias:
\begin{align}
\tilde{H} &= H_M + W_{\text{ori}} \hat{\bm{R}}^b + W_{\text{trans}} \hat{\bm{t}}^b, \\
H'' &= \mathrm{SelfAttn}(\tilde{H}), \\
C^{\text{PP}} &= \mathrm{softmax}\bigl(Q'' K_S^\top / \sqrt{d} + P^b\bigr) V_S, \\
H^{\text{PP}} &= \mathrm{MLP}\bigl([H'', C^{\text{PP}}]\bigr).
\end{align}
A final decoder assembles the predicted poses and trajectory into the SMPL-X future motion $\hat{Y}^b_{t+1:T}$.

\paragraph{Configuration.}
We stack $3$ trajectory blocks and $3$ pose blocks with hidden dimension $256$, $4$ attention heads, and dropout $0.1$.

\paragraph{Comparison to gaze-conditioned decoders.}
Gaze-conditioned scene-aware HMP models~\cite{lou2024sif3d, yu2025vision} inject a per-frame 3D gaze point as the scene-conditioning signal, which most directly affects the trajectory branch.
Our predictor instead consumes the selected token $i^*_b$, inferred from $(M_{1:t}, S)$ through the prior and requiring no auxiliary signal at inference; the token participates in both branches (its embedding in the trajectory branch and its mask through $P^b$ in the pose branch), so the inferred grounding shapes trajectory and pose jointly.

\subsection{Teacher Pose Refiner}
\label{app:refiner}

\paragraph{Architecture.}
The teacher-only pose refiner (Section~\ref{sec:decode}) adjusts the initial pose $\tilde{\bm{\theta}}^{\Tch}_{t+1:T}$ from $D^{\Tch}_{\text{pose}}$ using the textual intent $\mathcal{I}$ and the selected token's embedding:
\begin{equation}
\hat{\bm{\theta}}^{\Tch}_{t+1:T} = \mathrm{Refine}\bigl(\tilde{\bm{\theta}}^{\Tch}_{t+1:T},\, \mathcal{I},\, \bm{e}_{i^*_{\Tch}}\bigr).
\label{eq:refiner}
\end{equation}
To preserve the geometrically grounded trajectory, $\mathrm{Refine}$ keeps the predicted translation fixed and allows only a small, zero-initialized orientation residual $\Delta\bm{R}^{\Tch}_{t+1:T}$, composed as $\hat{\bm{R}}^{\Tch}_\tau \leftarrow \hat{\bm{R}}^{\Tch}_\tau \cdot \Delta\bm{R}^{\Tch}_\tau$ in $SO(3)$; this permits limited re-orientation (e.g., turning slightly toward an object before sitting) without disrupting the global path.
Since the student has no access to $\mathcal{I}$, $\mathrm{Refine}$ is teacher-only; the student's pose comes directly from $D^{\Stu}_{\text{pose}}$.

\paragraph{Intent-alignment loss.}
$\mathrm{Refine}$ is supervised by an intent-alignment loss that pulls the refined pose toward $\mathcal{I}$ in a shared text--pose feature space:
\begin{equation}
\mathcal{L}_{\text{intent-pose}} =
\lambda_{\text{end}}\,d\bigl(\psi_\theta(\hat{\bm{\theta}}^{\Tch}_T),\, \psi_c(\mathcal{I})\bigr)
+ \lambda_{\text{seq}}\,d\bigl(\psi_\theta(\hat{\bm{\theta}}^{\Tch}_{t+1:T}),\, \psi_c(\mathcal{I})\bigr),
\label{eq:intent_pose}
\end{equation}
where $\psi_c$ is a text encoder, $\psi_\theta$ a pose-sequence encoder mapping body pose into the same latent space, and $d(\cdot,\cdot)$ a cosine distance.
The first term enforces alignment at the destination frame, which corresponds to the interaction state described by $\mathcal{I}$, while the second propagates intent semantics across the sequence.
The orientation residual is regularized by $\lambda_{\text{ori}}\,\lVert\Delta\bm{R}^{\Tch}_{t+1:T}\rVert$, keeping orientation close to the trajectory anchor.
Values are listed in Table~\ref{tab:hparams}.

\FloatBarrier
\section{Training and Implementation Details}
\label{app:impl}
\label{app:hparams}

\subsection{Losses}
\label{app:losses}

\paragraph{Motion reconstruction loss.}
For branch $b \in \{\Tch, \Stu\}$,
\begin{equation}
\mathcal{L}^b_{\text{mot}} =
\lambda_{\text{joint}}\,\mathcal{L}_{\text{joint}}
+ \lambda_{\text{vp}}\,\mathcal{L}_{\text{vposer}}
+ \lambda_{\text{mpjpe}}\,\mathcal{L}_{\text{mpjpe}}
+ \lambda_{\text{vel}}\,\mathcal{L}_{\text{vel}}
+ \lambda_{\text{foot}}\,\mathcal{L}_{\text{foot}}
+ \lambda_{\text{sem}}\,\mathcal{L}_{\text{sem}},
\end{equation}
comprising joint-level reconstruction on 3D joint positions, VPoser-latent regularization toward the pose prior~\cite{pavlakos2019smplx}, non-root MPJPE, inter-frame velocity consistency, a foot-contact penalty that suppresses sliding during ground contact, and semantic-prompt alignment; individual terms follow~\cite{lou2024sif3d, yu2025vision}.

\paragraph{Optimized parameters.}
\begin{itemize}
    \item Teacher trajectory pass: updates $\{E^{\Tch}_M, E_{\mathcal{C}}^{\Tch}, D^{\Tch}\}$ with $\mathcal{L}^{\Tch}_{\text{mot}}$.
    \item Teacher refinement pass: freezes $\{E^{\Tch}_M, E_{\mathcal{C}}^{\Tch}, D^{\Tch}_{\text{traj}}\}$ and updates $\mathrm{Refine}$ (with $\Delta\bm{R}^{\Tch}$) using $\mathcal{L}_{\text{intent-pose}}$, pose reconstruction, and the residual penalty.
    \item Student: freezes the teacher and updates $\{E^{\Stu}_M, E_{\mathcal{C}}^{\Stu}, F^{\text{grd}}_{\Stu}, D^{\Stu}\}$ using $\mathcal{L}^{\Stu}$ (Eq.~\ref{eq:student_obj}), treating $\pi^{\Tch}$ and $\hat{Y}^{\Tch}_{t+1:T}$ as stop-gradient targets.
\end{itemize}

\subsection{Optimization and Inference}
\label{app:training}

The trajectory and pose branches are jointly optimized
with AdamW (initial learning rate $3\times10^{-4}$,
weight decay $10^{-4}$) and an exponential learning-rate decay
($\gamma=0.98$ per epoch), with batch size
8 and a training budget of up to 100 epochs.

At inference, the VLM, the LLM, textual intent, and the teacher are not used:
given only $M_{1:t}$ and $S$, the student predicts $\pi^{\Stu}$,
selects $i^*_{\Stu}$ (Eq.~\ref{eq:select}), and forecasts
\begin{equation}
\hat{Y}^{\Stu}_{t+1:T}
=
D^{\Stu}\bigl(
\bm{h}^{\Stu}_M,\,
S,\,
\bm{m}_{i^{*}_{\Stu}},\,
\bm{e}_{i^{*}_{\Stu}}
\bigr).
\end{equation}

\subsection{Hyperparameters}

\begin{table}[h]
\centering
\small
\caption{Hyperparameters used in all GIMO and PROX experiments.}
\label{tab:hparams}
\setlength{\tabcolsep}{5pt}
\begin{tabular}{lll}
\toprule
Group & Hyperparameter & Value \\
\midrule
Candidate tokens & cone aperture & $120^\circ$ \\
                 & GT--proposal matching IoU (evaluation) & $0.5$ \\
Teacher prior    & fusion weight $\alpha$, temperature $\tau$ (Eq.~\ref{eq:teacher_prior}) & $0.79$, $1.0$ \\
                 & $w_{\text{face}}, w_{\text{body}}, w_{\text{line}}$ (Eq.~\ref{eq:geo}) & $0.5, 2, 0.5$ \\
Gaze variant     & $\lambda_d, \lambda_a$ & $3, 3$ \\
Encoders         & $E_M$ & 4-layer Transformer, $d=256$, 4 heads \\
                 & $E_{\mathcal{C}}$ & PointNet++ (2 SA layers), $d_e = 16$ \\
                 & descriptor dimension $F$ & $16$ \\
Scoring head     & hidden size, MLPs & $256$; 2-layer MLPs, GELU \\
Motion predictor & trajectory / pose blocks & $3$ / $3$ \\
                 & hidden size, heads, dropout & $256$, $4$, $0.1$ \\
Student loss     & $\lambda_z$, $\lambda_Y$ (Eq.~\ref{eq:student_obj}) & $1.0$, $0.5$ \\
                 & $\lambda_{\text{traj}}$, $\lambda_{\text{pose}}$ (Eq.~\ref{eq:ly_kd}) & $1.0$, $0.5$ \\
Motion loss      & $\lambda_{\text{vel}}$ & $0.2$ \\
                 & $\lambda_{\text{vp}}, \lambda_{\text{mpjpe}}, \lambda_{\text{foot}}, \lambda_{\text{sem}}$ & $1.0$, $1.0$, $0.1$, $0.1$ \\
Pose refiner     & $\lambda_{\text{end}}$ & $0.1$ \\
                 & $\psi_c$ / $\psi_\theta$ MLP dims & $1536{\to}256{\to}256$ / $66{\to}256{\to}256$ \\
Optimization     & optimizer, lr, weight decay & AdamW, $3\times10^{-4}$, $10^{-4}$ \\
                 & schedule, batch size, epochs & exponential ($\gamma{=}0.98$), 8, 100 \\
\bottomrule
\end{tabular}
\end{table}

\subsection{Compute, Seeds, Assets, and Code}
\label{app:compute}

All experiments run on a single NVIDIA RTX~3090 (24\,GB).
Each training stage takes approximately 2 hours.
The VLM intent extraction and the LLM rubric scoring are run once per dataset and cached (the LLM pass over the GIMO training set takes approximately 4 hours); no VLM or LLM call occurs during training updates or inference.
SGIP results are averaged over five random seeds ({7, 42, 100, 984, 1004}).
We use GIMO~\cite{zheng2022gimo}, PROX~\cite{hassan2019prox}, HIK~\cite{tanke2023hik}, and HOI-M$^3$~\cite{zhang2024hoim3}, together with OpenIns3D~\cite{huang2024openins3d}, Qwen3~\cite{yang2025qwen3}, Qwen3-VL~\cite{bai2025qwen3vl}, and Idefics3~\cite{laurencon2024idefics3}, under their respective licenses and terms of use for non-commercial research.

\subsection{Third-Party Assets and Licenses}
\label{sec:asset-licenses}

\paragraph{Datasets.}
We use GIMO~\cite{zheng2022gimo}, PROX~\cite{hassan2019prox},
HIK~\cite{tanke2023hik}, and HOI-M$^3$~\cite{zhang2024hoim3}.
GIMO is distributed under a custom non-commercial research agreement
that restricts redistribution.
PROX is provided under the Software Copyright License for
non-commercial scientific research purposes.
HIK uses a custom non-commercial dataset license, specified in
Section 1.3 of its supplementary material.
The HIK API code is separately MIT-licensed.
For HOI-M$^3$, the official toolbox is Apache-2.0 licensed, but a
dataset-specific license was not identified on the public distribution
pages.
We do not attribute the toolbox license to the dataset.

\paragraph{Pretrained models.}
The official model cards declare Apache-2.0 for
Qwen3-14B~\cite{yang2025qwen3},
Qwen3-VL-8B-Instruct and Qwen3-VL-4B-Instruct~\cite{bai2025qwen3vl},
and Idefics3-8B-Llama3~\cite{laurencon2024idefics3}.
SMPL-X~\cite{pavlakos2019smplx} is distributed under the Software
Copyright License for non-commercial scientific research purposes.
The official VPoser code repository specifies the same license
title.

\paragraph{Implementations and distribution.}
The OpenIns3D implementation~\cite{huang2024openins3d} is
MIT-licensed.
Our PointNet++ backbone~\cite{qi2017pointnetpp} uses Erik Wijmans' Pointnet2\_PyTorch implementation, distributed under the Unlicense.
We also acknowledge the GIMO~\cite{zheng2022gimo} and
SIF3D~\cite{lou2024sif3d} implementations and documentation for
setup and data conventions.
Third-party assets remain subject to their original licenses and
terms of use. Raw datasets, processed dataset caches, and third-party
model weights are not included in the source release.

\FloatBarrier
\section{Broader Impacts}
\label{app:broader}

Anticipating where a person will move and how they will interact with the scene can make assistive robots and embodied agents safer and more responsive, and SGIP's observation-only deployment avoids gaze trackers or user-provided intent.
The same capability could, however, be used to track or anticipate people's behavior without their consent.
SGIP is trained on public research datasets of consenting participants and does not identify individuals, but deployments that observe people should obtain consent and comply with applicable privacy regulations.
Predictions can also differ from a person's actual choice in ambiguous scenes (Appendix~\ref{app:alternative}), so safety-critical systems should not rely on a single predicted future; the ranked alternatives of Section~\ref{sec:multifuture} can support more conservative planning.


\end{document}